# Diffusion-RL Based Air Traffic Conflict Detection and Resolution Method

Tonghe Li, Jixin Liu, Weili Zeng, Hao Jiang

*College of Civil Aviation, Nanjing University of Aeronautics & Astronautics, 211106*

**Abstract**

In the context of continuously rising global air traffic, efficient and safe Conflict Detection and Resolution (CD&R) is paramount for air traffic management. Although Deep Reinforcement Learning (DRL) offers a promising pathway for CD&R automation, existing approaches commonly suffer from a "unimodal bias" in their policies. This leads to a critical lack of decision-making flexibility when confronted with complex and dynamic constraints, often resulting in "decision deadlocks." To overcome this limitation, this paper pioneers the integration of diffusion probabilistic models into the safety-critical task of CD&R, proposing a novel autonomous conflict resolution framework named Diffusion-AC. Diverging from conventional methods that converge to a single optimal solution, our framework models its policy as a reverse denoising process guided by a value function, enabling it to generate a rich, high-quality, and multimodal action distribution. This core architecture is complemented by a Density-Progressive Safety Curriculum (DPSC), a training mechanism that ensures stable and efficient learning as the agent progresses from sparse to high-density traffic environments. Extensive simulation experiments demonstrate that the proposed method significantly outperforms a suite of state-of-the-art DRL benchmarks. Most critically, in the most challenging high-density scenarios, Diffusion-AC not only maintains a high success rate of 94.1% but also reduces the incidence of Near Mid-Air Collisions (NMACs) by approximately 59% compared to the next-best-performing baseline, significantly enhancing the system's safety margin. This performance leap stems from its unique multimodal decision-making capability, which allows the agent to flexibly switch to effective alternative maneuvers. This research not only provides a substantially more robust solution to the complex CD&R problem but also validates the immense potential of diffusion models as a highly expressive policy representation tool for safety-critical autonomous systems. By doing so, it opens up a new theoretical and practical pathway for the development of the next generation of intelligent air traffic management systems.

## 1 Introduction

In the context of continuously rising global air traffic volume, en route air traffic management (especially Area Control) is critical to the safety and efficiency of the entire air transportation system. Within expansive, densely trafficked area-control airspace, multiple aircraft cruise at high speed along different routes and flight levels, making potential conflicts the foremost risk factor affecting both safety and operational efficiency. Consequently, conflict detection and resolution (CD&R) in area control is not only the core responsibility of controllers to ensure flight safety and maintain traffic order, but also a key technology determining sector capacity and smoothness of operations. Timely and accurate detection of potential conflicts, coupled with the formulation and implementation of precise, efficient resolution strategies that minimally disturb normal operations, is of paramount importance. Faced with increasingly complex traffic patterns, shrinking separation minima and the continuous pursuit of higher safety standards, how to improve the foresight and reliability of conflict detection, and to optimize the intelligence, coordination and overall performance of resolution strategies, has become a critical scientific challenge and technical bottleneck in air traffic management—driving theoretical innovation and technological transformation.

To address these escalating CD&R challenges in area control and meet future ATM demands for greater safety margins, higher capacity and optimal operational efficiency, automation—and to some extent autonomy—of CD&R has emerged as a prominent research trend and frontier in international ATM. A core vision is the development of advanced CD&R capabilities under the trajectory-based operations (TBO) paradigm, enabling early conflict prediction based on precise four-dimensional (4D) trajectories and supporting more collaborative, optimized resolution schemes, as demonstrated and continuously evolved in modern ATM initiatives such as NextGen in the United States and SESAR in Europe [1–3]. Within this framework, leveraging artificial intelligence (AI), particularly deep reinforcement learning (DRL), to train agents that autonomously learn and execute conflict-resolution strategies has become a highly promising research focus. Recent studies have shown that DRL agents can acquire effective—or even superior—resolution maneuvers compared to traditional heuristics in complex multi-aircraft conflict scenarios without pre-specified rules, thereby potentially reducing controller workload and improving resolution efficiency [4–12]. Although these forward-looking directions still require extensive technical and safety validation before operational deployment, they have undeniably charted key pathways and laid a solid theoretical and technological foundation for tackling increasingly complex CD&R problems and building the next generation of intelligent, efficient

ATM systems.

Despite significant progress in automated and intelligent CD&R research, existing DRL approaches still face several key challenges when applied to the complex three-dimensional conflict-resolution tasks of area control, limiting their real-world performance. First, there is an inherent limitation in representing strategy diversity and decision flexibility. In 3D area-control environments, a single conflict can be resolved safely by qualitatively different maneuvers—such as heading change alone, flight-level change alone, or specific combinations thereof—but many mainstream DRL algorithms employ policy-network designs that tend to produce a single peaked action distribution. This unimodal bias makes it difficult for the agent to inherently retain and express multiple simultaneously valid resolution options. When several resolution paths are feasible, such representational insufficiency may cause the agent to fixate on the first or slightly highest-valued maneuver it learns, thereby lacking sufficient flexibility under dynamic constraints (for example, limited horizontal or vertical maneuvering space) and missing other equally valid or more suitable options. Second, the ability to learn efficiently and discover high-quality composite maneuvers in a complex decision space remains bottlenecked. The CD&R action space is high-dimensional and discrete—a factorized intention space over heading change, speed change, and flight-level step ($3\times3\times3 = 27$ classes)—and exploration mechanisms in many existing methods often lack sufficient coverage and efficiency over this joint discrete space. As a result, agents may struggle to identify non-intuitive but highly effective composite maneuvers (for example, particular combinations of heading, speed and flight-level changes) tailored to certain multi-aircraft conflicts. This exploration limitation can slow learning, degrade final resolution success rates and even prevent stable convergence in complex scenarios, hindering the development of robust decision-making capabilities. Moreover, how to induce agents to generate resolution commands that both guarantee separation and provide fine-grained, flexible maneuvers while minimizing disturbance to flight plans—rather than coarse, reactive avoidance actions—remains a major obstacle to practical CD&R systems. These limitations collectively define the research challenges that must be overcome to elevate agent decision quality and autonomous resolution performance in the specific context of 3D CD&R.

To address the above-mentioned limitations in current CD&R methods—namely the lack of strategy-diversity expression, insufficient exploration in complex 3-D environments and coarse resolution-command quality—we propose a novel deep-reinforcement-learning framework based on diffusion probability models for autonomous 3-D CD&R in area-control airspace. Unlike conventional actor networks that impose a unimodal (often Gaussian) prior on the policy, we formulate the policy itself as the reverse trajectory of a stochastic-differential-equation–driven diffusion process: the forward SDE progressively injects Gaussian noise into any action sample, while the reverse SDE—parameterized by a neural network that estimates the score (the gradient of the log-density)—iteratively removes this noise to regenerate actions. With a sufficiently accurate score estimator, the diffusion process can in principle approximate an arbitrary target action distribution to any desired precision. This capability allows the agent to represent rich multi-peak action distributions, thereby overcoming the inherent unimodal bias of traditional policies and enabling multiple qualitatively different yet safe and effective resolution maneuvers for the same conflict scenario (e.g., flexibly combining heading, speed and flight-level changes).Furthermore, by integrating diffusion-based policy learning into a customized Actor–Critic architecture and incorporating safety-oriented mechanisms—including dual-Q conservative evaluation and a density-progressive conflict curriculum—our method markedly improves exploration efficiency and training stability in high-dimensional action spaces. On the one hand, the **diffusion policy's inherent stochasticity** allows the agent to comprehensively explore the vast state–action space, mastering diverse and efficient resolution maneuvers and thus avoiding local optima; on the other hand, the iterative diffusion-based generation process yields finer-grained, operation-compliant commands rather than coarse reactive avoidance.

These methodological novelties translate into three concrete contributions, summarized below.

(1) We pioneer the application of a diffusion-probability-model-based DRL framework to the safety-critical and highly challenging problem of 3D CD&R in area control. We introduce a unified Diffusion-RL strategy for multi-aircraft 3D conflicts that directly generates multimodal resolution maneuvers. The action space is high-dimensional and discrete, instantiated as a factorized intention space over heading change, speed change, and flight-level step (3×3×3 = 27 classes), which enables diverse alternatives beyond traditional unimodal policies.

(2) We propose an online-learned customized Diffusion-Actor–Critic architecture for safety-critical tasks, incorporating dual Q-critic conservative evaluation to yield an online-learned, highly safe and robust Diffusion-Policy. This marks the first engineering-scale deployment of the Diffusion-AC framework in an ATC safety scenario.

(3) We introduce a Density-Progressive Safety Curriculum (DPSC) that starts with low-density

conflicts under relaxed penalties and progressively tightens both traffic density and safety weights, reducing convergence steps by ≈ 14 % and NMAC incidence by ≈ 59 % under peak traffic conditions. This is the first curriculum-based training scheme tailored to diffusion-model CD&R, significantly boosting sample efficiency and safety robustness.

## 2 Literature Review

Traditional research in Conflict Detection and Resolution (CD&R) has primarily revolved around the paradigm of "early prediction and early resolution," leveraging trajectory prediction alongside geometric and optimization models for conflict detection and maneuver planning, with operational viability being progressively validated in procedural ATC. The seminal review by Kuchar and Yang provided a systematic categorization of these modeling schools, establishing a foundational terminology and evaluation framework [13]. In the pursuit of provably safe solutions, Tomlin et al. formulated conflict avoidance as a problem of hybrid systems and reachable sets, enabling the synthesis of maneuvers with strict safety guarantees[14,15]. Along an engineering-oriented track, Erzberger's team developed integrated automated resolution and arrival management systems for separation assurance, demonstrating their robustness via large-scale fast-time simulations in realistic, high-demand scenarios[16–18]. Meanwhile, another school of thought has focused on optimization, using mixed-integer programming to address the coupled challenges of multi-aircraft conflict avoidance and sequencing, with an emphasis on global consistency and interpretability[19]. In the domain of distributed systems, the ORCA framework for reciprocal collision avoidance established the necessary and sufficient conditions for agents to share responsibilities, becoming a cornerstone of multi-agent collision avoidance research [20]. Overarching these developments, Trajectory-Based Operations (TBO) has emerged as the main thrust in modern ATM evolution, fostering a new generation of tools for separation management by emphasizing a planning-execution loop driven by high-precision 4D trajectories. These established lines of research have collectively defined the baseline requirements for any solution—it must be "predictable, interpretable, and verifiable." However, they also revealed the practical limits of their methodologies: when faced with escalating situational complexity, traffic density, and constraint dimensions, the costs of purely geometric or optimization-based methods in terms of real-time performance and strategic diversity become prohibitively high, particularly in complex conflicts that admit multiple, equally viable resolutions.

Recent research into learning-based and Reinforcement Learning (RL) for CD&R has converged on several key trends: the adoption of multi-agent systems, the development of end-to-end perception-to-decision pipelines, and a growing integration with the broader ecosystem of simulation tools. One prominent line of research leverages open-source simulators to establish high-density airspace benchmarks for exploring both centralized and distributed multi-aircraft separation strategies. Studies in this area consistently report that learned policies achieve more stable separation assurance and superior delay control compared to traditional heuristics in canonical scenarios such as head-on, crossing, and overtaking encounters [21–24]. Notably, Hoekstra and Ellerbroek introduced the BlueSky ATC simulator, a fully open-source, agent-based platform for air traffic simulation [21]. Building on this, Groot et al. developed the BlueSky-Gym API, which integrates a suite of RL environments based on Gymnasium and BlueSky, thereby providing a standardized framework for RL research in air traffic management[22]. In a similar vein, Brittain et al. proposed a novel deep multi-agent reinforcement learning (MARL) framework to identify and resolve conflicts among a variable number of aircraft in high-density airspace. Their research realized an autonomous separation assurance function that provides safe spacing in congested en-route sectors, addressing the limitations of conventional ATC systems in complex scenarios [23]. Subsequently, the same group developed a decentralized RL framework that incorporates attention networks to deliver autonomous separation capabilities for advanced air traffic corridors. This method is particularly tailored for highly automated air traffic environments, enhancing flight safety and operational efficiency via a distributed learning mechanism and addressing the real-time decision-making challenges inherent in large-scale unmanned aircraft systems (UAS) traffic management [24].

A second category of research prioritizes interpretability and controllability, aiming to balance policy quality with effective human-machine collaboration by integrating mechanisms such as graph network attention, hierarchical policies, or rule-based priors into the training process[6–11,25–30]. For instance, Chen et al. developed general multi-agent reinforcement learning (MARL) frameworks with adaptive maneuver strategies for real-time conflict resolution in both 2D and 3D environments[6,7]. Efforts to enhance transparency are exemplified by Wang et al., who designed a DRL controller that decouples Q-value learning into distinct safety and efficiency modules to improve interpretability and decision support [9]. Sui et al. first applied Independent Deep Q-Networks (IDQN) to the problem, achieving strategies with high operational fidelity [10], and later extended this work to tactical resolution in three-dimensional action spaces [31].

Graph Neural Networks (GNNs) have also been a key focus. Li et al. developed a Graph Reinforcement Learning (GRL) method to efficiently handle the increasing airspace density [11], while Vouros et al. used graph-convolutional RL within their ResoLver system to enhance both automation and operational transparency, considering the interests of both controllers and airspace users [25]. A novel approach by Zhao and Liu involves injecting prior physical knowledge into the DRL algorithm, enabling the neural network to learn more efficiently from incomplete data and improving resolution accuracy [30]. Other significant MARL contributions include the Joint-CRAS framework by Huang et al. for simultaneously managing multiple conflicts [28] and the work of Isufaj et al. on modeling pairwise conflicts to improve resolution quality [29]. Complementing these algorithmic advances, Groot et al. systematically analyzed the critical impact of traffic density on the training of RL-based methods [8], while a comprehensive survey by Wang et al. reviewed the fundamental principles, evolution, and application landscape of deep reinforcement learning in air traffic conflict resolution [27].

Research in settings that more closely approximate operational realities has evaluated the impact of variables such as traffic density, formation size, and specific task contexts (e.g., Urban Air Mobility or ground taxiing) on training stability and policy generalization. These studies often attempt to suppress the risks of abnormal maneuvers and Near Mid-Air Collisions (NMACs), which can arise from "greedy" solution-seeking, by incorporating explicit safety constraints or sophisticated reward shaping techniques [32–36]. For instance, Nilsson et al. developed a self-prioritizing multi-agent reinforcement learning method to resolve conflicts under the constraint of limited instructions [32], while Han and Huang proposed an RL-based model for flight separation assurance designed to overcome the shortcomings of existing methods in complex airspace [33].

Nevertheless, a fundamental bottleneck persists: mainstream deep RL policies typically rely on Gaussian or other unimodal approximations. This architectural choice makes it exceedingly difficult to simultaneously represent and maintain multiple, equally viable resolution patterns—such as specific combinations of composite maneuvers—within a three-dimensional discrete action space. Consequently, under dynamic constraints, these approaches often exhibit policy rigidity and insufficient exploration, a critical limitation that this work aims to address.

Policy learning with diffusion models offers a methodological breakthrough to this bottleneck. By characterizing the action distribution's "gradient field" through a conditional denoising process, diffusion policies are inherently adept at representing the multimodal, high-dimensional, and temporally extended distributions required for complex strategies. Their superior stability and performance have been repeatedly demonstrated on robotics and offline RL benchmarks [37–43]. Furthermore, the integration of value or policy guidance introduces a gradient-based or re-weighting mechanism that biases the diffusion process toward high-value regions. This technique enhances alignment with task rewards while preserving crucial distributional diversity, as evidenced by the state-of-the-art or highly competitive results achieved by methods like Diffusion-QL, Policy-Guided Diffusion, QVPO, and DreamFuser on benchmarks such as D4RL [38,42–46].

This approach is fundamentally aligned with the "value guidance" principle introduced in Section 4 of this paper. We leverage a critic (Q-function) to weight, filter, and distill the denoised logits or candidate intentions, thereby forming a differentiable value-consistency constraint without causing the policy head to degenerate back into a unimodal approximation. This is complemented by a Density-Progressive Safety Curriculum (DPSC), a custom training scheme tailored for our Diffusion-AC architecture. Through a linear tightening of stage-advancement gates and Loss of Separation (LoS) thresholds, we facilitate a gradual training convergence, progressing from sparse to dense traffic and from relaxed to strict safety criteria. Crucially, this curriculum only modifies scenario density, termination thresholds, and reward weights; the action feasibility mask remains invariant throughout, ensuring the consistency of the policy-safety interface and yielding provable gains in training stability. From the broader perspective of TBO practice, this combination of a "multimodal policy + value guidance + curriculum learning" synergistically balances the expressive power needed to capture multiple, equally optimal solutions with the safe, operationally-oriented convergence required for deployment. This synergy offers a plausible pathway for the operational implementation of 3D-CD&R in high-density airspace.

## 3 Methodology

### 3.1 Problem Description

During Air Traffic Management (ATM) operations, the core duties of an Area Control controller fall into three categories. First, controllers accept and hand over flights, ensuring that each aircraft operates safely within their sector in accordance with its flight plan—typically entering the sector from a designated transfer point at a specified flight level and following the prescribed route to the next transfer point. Second, they continuously monitor the airspace to detect and anticipate potential conflicts between flights in a timely manner, such as head-on, overtaking, or converging situations. Finally—and most critically—they resolve these potential conflicts by issuing control instructions. In practice, controllers can employ various combinations of measures, including heading adjustments (e.g., direct routing, radar vectoring, or lateral offsets), speed control, and flight-level allocation.

According to the fundamentals of reinforcement learning—namely, the Markov process, in which every time step comprises a state and an action—the conflict-detection-and-resolution task can be decomposed accordingly: the state (i.e., the observation) accounts for conflict detection, while the action addresses conflict resolution. In this study, we train an aircraft-mounted deep-reinforcement-learning agent based on a diffusion probabilistic model. At each time step, the agent observes its own state together with information on the three nearest aircraft. It then selects actions along three dimensions—heading adjustment, speed adjustment, and flight-level adjustment—to avoid conflicts with intruders and ultimately reach the designated waypoint. **Fig. 1** illustrates the air-traffic conflict-detection-and-resolution task.

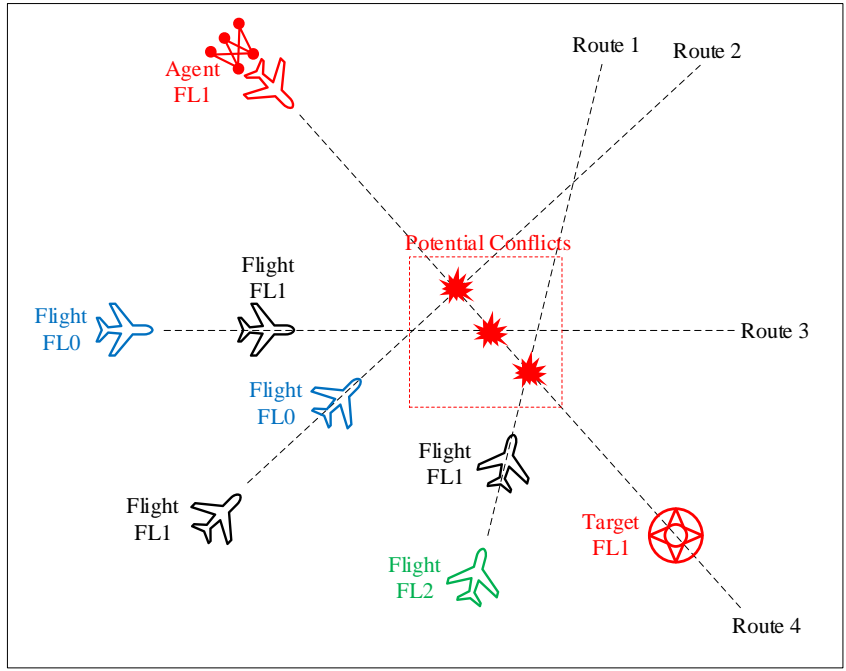

**Fig. 1 Illustration of air-traffic conflict detection and resolution**

## 3.2 Model Formulation

### 3.2.1 Mathematic Representation

Air-traffic conflict detection and resolution is, in essence, a constrained multi-objective optimization problem. Let the airspace $\Omega \subset \mathbb{R}^2 \times \mathbb{Z}^+$ be a bounded discrete domain whose horizontal dimension is a continuous plane and whose vertical dimension consists of discrete flight-level strata. The airspace contains N aircrafts, and the complete state vector of aircraft $f_i$ at time t is denoted as:

$$\mathbf{s}_i(t) = [x_i(t), y_i(t), v_{x,i}(t), v_{y,i}(t), \psi_i(t), v_i(t), FL_i(t)]^T \in \mathbb{R}^6 \times \mathbb{Z}^+ \quad (1)$$

where $(x_i, y_i)$ denotes the aircraft's horizontal position in the airspace, $(v_{x,i}, v_{y,i})$ denotes the horizontal-velocity vector component, $\psi_i$ is the heading angle (in radians), $v_i$ is the scalar airspeed, $FL_i \in \{0,1,2\}$ is the index of the discrete flight-level stratum, each level corresponding to an actual altitude increment of 300 m. For the target aircraft (the controlled agent), its kinematic model follows the standard aircraft flight dynamics:

$$\begin{cases} \dot{x}_{agent}(t) = v_{agent}(t)\cos(\psi_{agent}(t)) \\ \dot{y}_{agent}(t) = v_{agent}(t)\sin(\psi_{agent}(t)) \\ \dot{\psi}_{agent}(t) = u_{\psi}(t) \\ \dot{v}_{agent}(t) = u_v(t) \\ FL_{agent}(t+1) = FL_{agent}(t) + u_{FL}(t) \end{cases} \quad (2)$$

where $\dot{x}_{agent}(t)$, $\dot{y}_{agent}(t)$ represent the rate of change of the agent's horizontal position at time t, respectively, $v_{agent}(t)$ is a scalar of the agent's flight velocity, $\psi_{agent}(t)$ is the heading angle (in radians), $\dot{\psi}_{agent}(t)$, $\dot{v}_{agent}(t)$ are the rate of change of heading angle and velocity respectively. $FL_{agent}(t)$ is the flight altitude layer index of the agent at time t. For the 3D control inputs: $u_{\psi}(t) \in \{-5°, 0°, 5°\}$ is the heading control input, $u_v(t) \in \{-50, 0, +50\}kt$ is the velocity control input, $u_{FL}(t) \in \{-1, 0, +1\}$ is the flight level control input. All the three commands are discrete.

### 3.2.2 Multi-objective optimization function

The core challenge of air traffic conflict detection and detection is to meet the requirements of flight efficiency and safety at the same time. In air traffic control practice, this is manifested in the need for controllers to minimize disruption to the normal operation of flights while ensuring spacing standards. Therefore, the system objective function is designed as two sub-objectives with definite physical significance:

1) Flight efficiency objective

The fundamental reason for setting flight efficiency objective is that one of the core missions of air traffic management is to ensure that flights operate punctually and efficiently. In real-world air traffic control operations, controllers must consider the following key factors when resolving conflicts:

first, flight delays will cause huge economic losses, including increased fuel costs, extended crew hours, passenger time costs, etc.; Secondly, deviation from the predetermined route will increase the flight distance and time, and reduce the overall airspace utilization efficiency. Third, excessive regulatory intervention can affect pilot operating load and passenger comfort; Finally, in high-density airspace, inefficient conflict resolution may trigger cascading delays and affect the operational order of the entire airspace. Therefore, the agent must minimize the deviation from the normal flight trajectory as much as possible and reach the target waypoint quickly and efficiently under the premise of ensuring safety.

The objective function quantifies the flight efficiency gains of the agent in the form of maximization:

$$J_G = \alpha_{goal} \cdot \mathbf{1}_{goal}(T) - \int_0^T \left[ \alpha_1 \| \mathbf{p}_{agent}(t) - \mathbf{p}_{goal} \|^2 + \alpha_2 \mathbf{1}_{boundary}(t) + \alpha_{step} \right] dt \tag{3}$$

where $J_G$ is the flight efficiency objective function, $T$ denotes the total time of the task, $\mathbf{p}_{agent}(t)$ denotes the agent's positive vector at time $t$, $\mathbf{p}_{goal}$ represents target waypoint location. Goal achievement rewards $\alpha_{goal} \cdot \mathbf{1}_{goal}(T) = 1.0 \times \mathbf{1}_{goal}(T)$ is a one-time positive reward for reaching the goal, where $\mathbf{1}_{goal}(T)$ is goal achievement indication function. The reward for achieving the goal is 1 if the goal is reached at the end of the task, otherwise it is 0. The integral term represents cumulative penalties, including a position-deviation penalty $\alpha_1 \| \mathbf{p}_{agent}(t) - \mathbf{p}_{goal} \|^2$ ($\alpha_1$=1), a boundary penalty $\alpha_2 \mathbf{1}_{boundary}(t)$ ($\alpha_2$=0.5), and a per–time-step penalty $\alpha_{step}$=0.01。$\mathbf{1}_{boundary}(t)$ is the boundary-indicator function, which equals 1 when the agent exits the airspace boundary and 0 otherwise.

2) Flight Safety Objective

The establishment of the flight safety objective is an absolute requirement and legal obligation in Air Traffic Management. The International Civil Aviation Organization (ICAO) places aviation safety at the highest priority of the civil aviation industry, and any conflict-resolution scheme must strictly comply with prescribed separation standards. In actual operation, violations of minimum separation can have severe consequences: at best, triggering ATC system alerts and demanding immediate corrective action by controllers; at worst, precipitating a mid-air collision with catastrophic losses. Moreover, near-mid-air collision (NMAC) events not only endanger flight safety but also provoke in-depth investigations by aviation authorities and potential legal liability. Modern ATC systems employ a layered safety-protection philosophy—through procedural separation, radar surveillance, conflict alerts, TCAS, and other measures—to ensure safety. Consequently, the agent's decision-making must place safety first, rigorously avoiding any behavior that could lead to a loss of separation or an NMAC.

This objective function quantifies the agent's reward for maintaining flight safety. It is formulated as a maximization problem, achieving the safety objective by penalizing conflict behaviors:

$$J_S = -\int_0^T \sum_{k=1}^{N} \left[ \mathcal{P}_{NMAC}(d_{agent,k}(t), FL_{agent,k}(t)) + \mathcal{P}_{LoS}(d_{agent,k}(t), FL_{agent,k}(t)) \right] dt \quad (4)$$

where $J_s$ is the flight–safety objective function, $N$ is the total number of intruders in the airspace, and $k$ indexes each of those aircraft. The horizontal distance between the agent and the $k$-th aircraft is $d_{agent,k}(t) = \| \mathbf{p}_{agent}(t) - \mathbf{p}_k(t) \|$, and their flight-level difference is $FL_{agent,k}(t) = |FL_{agent}(t) - FL_k(t)|$. $\mathcal{P}_{NMAC}(\cdot)$ and $\mathcal{P}_{LoS}(\cdot)$ denote the Near Mid-Air Collision penalty function and the Loss of Separation penalty function, respectively.

The Near Mid-Air Collision (NMAC) penalty function is defined as:

$$\mathcal{P}_{NMAC}(d, FL_{diff}) = \begin{cases} 1, & \text{if } d < d_{NMAC} \text{ and } FL_{diff} = 0 \\ 0, & \text{otherwise} \end{cases} \quad (5)$$

where d is the horizontal distance between the agent and an intruder, $FL_{diff}$ denotes the flight-level difference, $d_{NMAC} = 0.2\text{km}$ is the NMAC critical distance. When two aircraft occupy the same flight level ($FL_{diff} = 0$) and their horizontal separation falls below $d_{NMAC}$, the function outputs a positive penalty of 1.0. Combined with the negative sign in front of the integral in the safety objective, this contributes -1.0 to the objective; otherwise, no penalty is applied.

The Loss of Separation (LoS) penalty function is defined as:

$$\mathcal{P}_{LOS}(d, FL_{diff}) = \begin{cases} 0.5, & \text{if } d_{NMAC} \leq d < d_{LoS} \text{ and } FL_{diff} = 0 \\ 0, & \text{otherwise} \end{cases} \quad (6)$$

where $d_{LoS}$ is the standard horizontal separation distance. If two aircraft are at the same flight level and their horizontal distance lies between $d_{NMAC}$ and $d_{LoS}$, the function returns a positive penalty of 0.5. With the negative sign in the objective, this yields a -0.5 contribution, indicating a separation violation that has not yet reached the NMAC threshold.

3) Composite Objective Function

The necessity of a composite objective arises from the inherent conflict and balancing requirement between safety and efficiency in air traffic management. In real-world operations, an absolute safety strategy (e.g., dramatically increasing separation minima or significantly reducing airspeed) would severely degrade airspace capacity and operational efficiency; whereas an overly aggressive efficiency strategy could elevate safety risks. In daily practice, controllers must seek an optimal trade-off between these two goals—ensuring compliance with mandated safety standards while minimizing interference with flight operations.

The composite objective is constructed as an equally weighted sum of the individual objectives:

$$\max J_{total} = J_G + J_S \tag{7}$$

### 3.2.3 Model Constrains

1) Agent Performance Constraints

The imposition of agent performance constraints is grounded in the physical limits and flight-dynamics characteristics of real aircraft. In practice, civilian aircraft are subject to multiple physical restrictions—such as aerodynamic performance, engine thrust, and structural strength—that bound their behavior: speed is limited by stall speed and maximum operating speed; heading changes are constrained by turn radius and load factor; and altitude changes are limited by maximum climb and descent rates. Ignoring these limits would yield maneuver commands that are infeasible in a real environment. Moreover, excessively aggressive maneuvers can degrade passenger comfort, accelerate airframe fatigue, and increase pilot workload. International civil-aviation standards (e.g., ICAO Annex 6) explicitly prescribe maneuver limits for commercial transport flights. Therefore, the agent's action space must be strictly confined within the aircraft's actual performance envelope to ensure that any generated resolution commands are truly executable.

This constraint can be formulated as follows:

$$\begin{cases} V_{min} \leq v_{agent}(t) \leq V_{max} \\ |\Delta \psi_{agent}| \leq \Delta \psi_{max} \\ |\Delta v_{agent}| \leq \Delta v_{max} \\ FL_{agent}(t) \in \{0,1,2\} \\ |\Delta FL_{agent}| \leq 1 \end{cases} \tag{8}$$

where $v_{agent}(t)$ denotes the agent's airspeed at time t, $V_{min}$ and $V_{max}$ are the minimum and maximum allowable airspeeds, respectively. $\Delta \psi_{agent}$ is the per-step change in heading, with $\Delta \psi_{max} = 5°$ as the maximum permitted adjustment per time step. $\Delta v_{agent}$ is the per-step change in speed, and $\Delta v_{max} = 50kt$ is the maximum allowable speed adjustment per step. $FL_{agent}(t)$ is the agent's flight-level index, constrained to the three discrete levels $\{0,1,2\}$. $\Delta FL_{agent}$ is the single-step change in flight level, and at most one level change is permitted per time step.

2) Separation-Standard Constraint

According to Area Control operating standards, the horizontal separation between any two aircraft occupying the same flight level must be at least 10 km. This constraint can be written as:

$$d_{agent,k}(t) \geq D_{sep}, FL_{agent}(t) = FL_k(t), \forall k \in \{1,2,...,N\} \tag{9}$$

where k indexes each of the other N intruders in the airspace, $d_{agent,k}(t)$ is the horizontal distance between the agent and aircraft k at time t. $D_{sep}$ is the prescribed standard horizontal separation. The constraint stipulates that the minimum-separation requirement applies only when the agent and another intruder are at the same flight level ($FL_{agent}(t) = FL_k(t)$), if they occupy different flight levels, vertical separation alone suffices and the horizontal-separation condition need not be enforced.

3) Airspace boundary constraints

During area control operations, an aircraft is only allowed to fly within the current sector until it is handed over to the next one, and may not cross the latitude, longitude or flight-level boundaries of the current sector. This constraint can be expressed as follows:

$$0 \leq x_{agent}(t) \leq W \tag{10}$$

$$0 \leq y_{agent}(t) \leq H \tag{11}$$

$$FL_{agent}(t) \in \{0,1,FL_{max}\} \tag{12}$$

where $x_{agent}(t)$ and $y_{agent}(t)$ are the agent's horizontal coordinates at time t; W and H are the sector's horizontal width and vertical height limits. $FL_{agent}(t)$ is the agent's flight-level index, with $FL_{max} = 2$, as the highest level, so the agent's vertical movement is confined to the discrete levels 0, 1 and 2. These constraints ensure that the agent always operates within the designated control sector, avoiding handover and coordination issues caused by boundary breaches.

### 3.3 Markov Decision Process Modeling

As shown in **Fig. 2**, in the conflict-detection-and-resolution process, an aircraft must observe its environment and take an action at each time step. This can be cast as a real-time decision-making problem and then formulated as a Markov decision process, providing the foundation for subsequent reinforcement-learning methods.

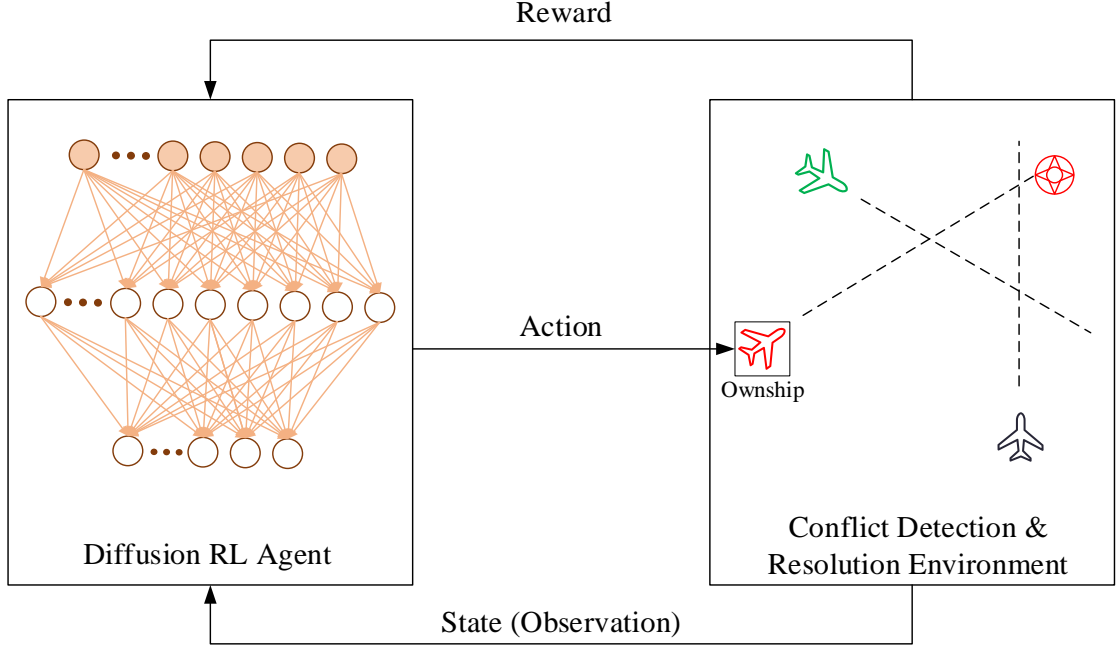

Fig. 2 Markov Decision Process Elements Mapping Diagram

**3.3.1 Definition of the MDP Five-Tuple**

We model the conflict-detection-and-resolution problem exactly as a Markov decision process (MDP), represented by the five-tuple:

$$\text{MDP} = \langle \mathcal{S}, \mathcal{A}, \mathcal{P}, \mathcal{R}, \gamma \rangle \tag{13}$$

where $\mathcal{S}$ is the state space, $\mathcal{A}$ the action space, $\mathcal{P}$ the state-transition probability function, $\mathcal{R}$ the reward function, and $\gamma$ the discount factor. This five-tuple fully specifies the MDP framework for our problem.

1) State space $\mathcal{S}$

The state space consists of the agent's observation vector, which corresponds exactly to the information the agent can observe:

$$\mathcal{S} = \{\mathbf{o}(t) : \mathbf{o}(t) = [\mathbf{o}_{intruders}(t), \mathbf{o}_{agent}(t), \mathbf{o}_{goal}(t)]\} \tag{14}$$

where **o**(t) is the observation vector at time t, **o**$_{intruders}$(t) is the state information of the nearest other aircraft observable by the agent, **o**$_{agent}$(t) is the agent's own state information, and **o**$_{goal}$(t) is the relative position of the target waypoint, given by:

$$\mathbf{o}_{goal}(t) = [(x_{agent}(t) - x_{goal}), (y_{agent}(t) - y_{goal})] \tag{15}$$

where $x_{goal}$ and $y_{goal}$ are the horizontal coordinates of the target waypoint. Designing $\mathbf{o}_{goal}(t)$ in this way ensures that the agent continuously senses its distance from the waypoint and is guided toward it. Meanwhile, $\mathbf{o}_{intruders}(t)$ provides information about intruders, directing the agent to detect potential conflicts.

2) Action space $\mathcal{A}$

To accommodate both fine-grained maneuvering and airline-standard flight-level assignments, we remodel the agent's action space as a factorized three-dimensional discrete space. This yields 3×3×3 = 27 discrete actions per step:

$$\mathcal{A} = \{-1, 0, +1\}_{\Delta\psi} \times \{-1, 0, +1\}_{\Delta v} \times \{-1, 0, +1\}_{\Delta h} \tag{16}$$

where the intention triple selects fixed-rate increments for heading, speed, and flight level:

$$\begin{cases} \Delta\psi_t \in \{-\Delta\psi_{step}, 0, +\Delta\psi_{step}\} \text{ (deg)} \\ \Delta v_t \in \{-\Delta v_{step}, 0, +\Delta v_{step}\} \text{ (kt)} \\ \Delta h_t \in \{-1, 0, +1\} \end{cases} \tag{17}$$

For the regional en-route scenarios studied in this article, we set

$$\begin{cases} \Delta\psi_{step} = 5° \\ \Delta v_{step} = 50 kt \end{cases} \tag{18}$$

which remain within typical ATC-approved maneuver limits while providing sufficient control authority for conflict resolution.

At each decision step $t$, the discrete action

$$a_t = (a_t^\psi, a_t^v, a_t^h) \in \{-1, 0, +1\}^3 \mapsto (\Delta\psi_t, \Delta v_t, \Delta h_t) = \left(a_t^\psi \Delta\psi_{step}, a_t^v \Delta v_{step}, a_t^h\right) \tag{19}$$

The action is applied to the controlled aircraft as

$$\begin{cases} \psi_{t+1} = \psi_t + \Delta\psi_t \\ v_{t+1} = \text{clip}(v_t + \Delta v_t, v_{min}, v_{max}) \\ h_{t+1} = h_t + \Delta h_t \end{cases} \tag{20}$$

where clip(·) enforces operational bounds on speed. This factorized discrete formulation lets the diffusion policy model multimodal distributions over the 27 joint actions while naturally respecting the integer flight-level constraint; the fixed step sizes ensure rate-limited maneuvers consistent with operational practice.

3) State-transition probability $\mathcal{P}$

Under the deterministic-environment assumption, if at time t the agent is in state $s_t$ and takes action $a_t$, it will deterministically transition to a unique next state $s_{t+1}$:

$$\mathcal{P}(s_{t+1} \mid s_t, a_t) = \begin{cases} 1, & \text{if } s_{t+1} = f(s_t, a_t) \\ 0, & \text{otherwise} \end{cases} \tag{21}$$

where $\mathcal{P}(s_{t+1} \mid s_t, a_t)$ gives the probability of moving to $s_{t+1}$ when action $a_t$ is executed in $s_t$. The function $f(s_t, a_t)$ is the deterministic state-transition mapping defined by the aircraft kinematic model and the environment update rules. Because the environment is assumed deterministic, the probability is either 1 (the transition must occur) or 0 (the transition cannot occur).

4) Reward function $\mathcal{R}$

The mathematical model's overall objective computes the total feedback the agent receives from the environment, but within the MDP framework the reward function must assign that feedback to each individual time step. To do this, we decompose the integral-form objective into per-step rewards. Specifically, at each time step the reward splits into a goal-achievement component and a safety component, defined (with signs unified to match the objective) as:

$$R_G(s_t, a_t) = \alpha_{goal} \mathbf{1}_{goal}(t) - \left[ \alpha_1 \| \mathbf{p}_{agent}(t) - \mathbf{p}_{goal} \|^2 + \alpha_2 \mathbf{1}_{boundary}(t) + \alpha_{step} \right] \tag{22}$$

$$R_S(s_t, a_t) = -\sum_{k=1}^{N} \left[ \mathcal{P}_{NMAC}\left(d_{agent,k}(t), FL_{agent,k}(t)\right) + \mathcal{P}_{LOS}\left(d_{agent,k}(t), FL_{agent,k}(t)\right) \right] \tag{23}$$

The total reward at each time step is then:

$$\mathcal{R}(s_t, a_t) = R_G(s_t, a_t) + R_S(s_t, a_t) \tag{24}$$

5) Discount factor $\gamma$

In this work we set $\gamma = 0.99$ to balance the importance of immediate versus long-term rewards. A relatively high gamma means the system values achieving long-term objectives more highly, ensuring that while the aircraft agent pursues instant safety, it does not overlook the ultimate goal of reaching the target waypoint.

**3.3.2 Softening of Constraints**

Because opening up a larger exploration space can help the agent learn more comprehensive and robust policies [47,48], we soften most of the model constraints defined in section 3.2.3—namely, constraints (9), (10) and (11). And we retain only constraint (8) to ensure that the agent respects the physical laws of aircraft operation, plus constraint (12) to keep it within the preset flight-level boundaries. In this way, during early exploration the agent may temporarily violate separation minima or even fly outside the sector boundary, but the steadily growing negative reward at each time step for such behavior will strongly guide the agent to avoid these violations and learn a more effective strategy for reaching the target waypoint.

**3.3.3 Policy Optimization Problem Modeling**

After the MDP five-tuple has been defined, the original continuous-time, multi-objective constrained optimization problem can be equivalently transformed into a discrete-time policy optimization problem:

$$\max_{\pi} \mathbb{E}_{\tau \sim \pi} \left[ \sum_{t=0}^{T} \gamma^t \mathcal{R}(s_t, a_t) \right] \tag{25}$$

where π is the policy function, τ is the state–action trajectory, $\mathbb{E}_{\tau \sim \pi}[\cdot]$ denotes expectation over trajectories under π, $T$ is the task horizon, $\gamma^t$ the discount factor at step t, and $\mathcal{R}(s_t, a_t)$ is the immediate reward at step $t$. The objective is to find the optimal policy $\pi^*$ that maximizes the expected cumulative discounted reward.

The constraints defining this policy optimization problem are:

$$\begin{cases} s_{t+1} = f(s_t, a_t) \\ a_t \in \mathcal{A}(s_t) \\ s_0 \sim \rho_0 \\ \mathcal{C}(s_t, a_t) \leq 0 \end{cases} \tag{26}$$

where $s_{t+1} = f(s_t, a_t)$ is the state-transition equation; $\mathcal{A}(s_t)$ is the admissible action set in state $s_t$; $s_0 \sim \rho_0$ indicates that the initial state $s_0$ follows the distribution $\rho_0$; and $\mathcal{C}(s_t, a_t) \leq 0$ is the set of constraint conditions—here limited to the aircraft kinematic constraints and flight-level boundary constraints. The policy $\pi : \mathcal{S} \to \Delta(\mathcal{A})$ maps each state to a distribution over actions, and the trajectory $\tau = (s_0, a_0, s_1, a_1, ...)$ denotes the sequence of state–action pairs.

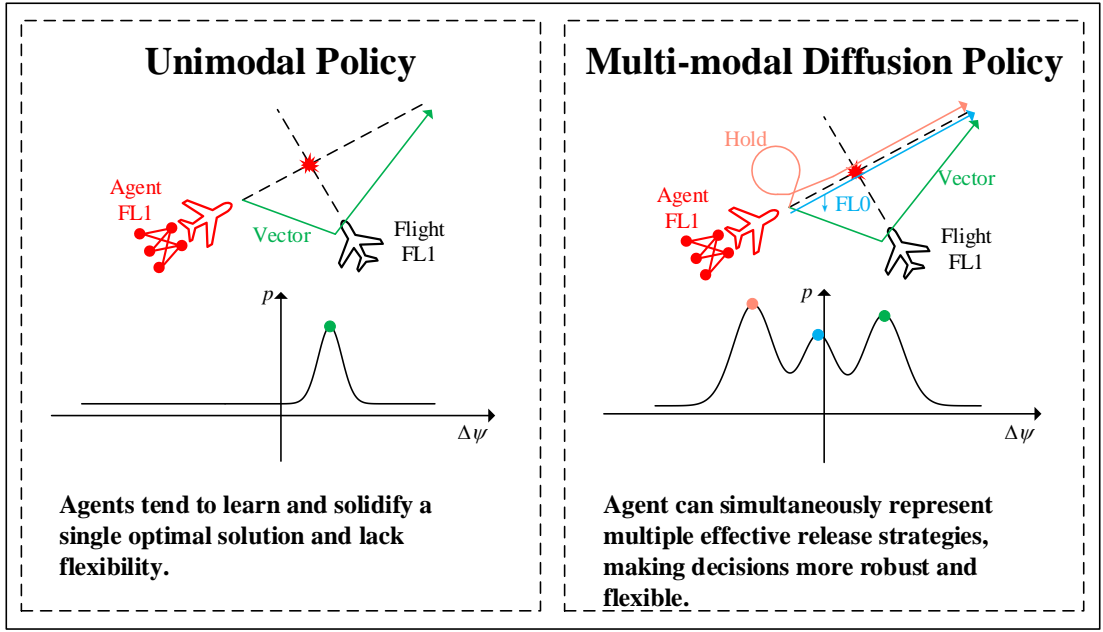

Fig. 3 Schematic Diagram of Core Idea Comparison

## 4 Algorithm Design

### 4.1 Motivation for a Diffusion Policy in Air Traffic CD&R

In regional air-traffic conflict-detection-and-resolution (CD&R), an agent must generate complex three-dimensional maneuvers—heading, speed and altitude changes—to maintain safe separation. As illustrated in **Fig. 3**, conventional deep-reinforcement-learning (DRL) algorithms such as TD3 or PPO typically parameterize the policy as a unimodal Gaussian. This single-peak assumption narrows the range of admissible actions: the agent is implicitly nudged toward one "dominant" resolution pattern and may fail whenever alternative maneuvers would be equally—or even more—effective. The limitation surfaced in Section 1 as a core obstacle to reliable CD&R.

#### 4.1.1 From Unimodal Bias to Multimodal Expressiveness

To overcome that bias we formulate the policy itself as the reverse trajectory of a stochastic-differential-equation (SDE)–driven diffusion process. In the forward SDE, Gaussian noise is gradually injected into any action sample until it becomes pure noise; in the reverse SDE, a neural network estimates the score function—the gradient of the log-density—and iteratively "denoises" the sample back to an action. With a sufficiently accurate score estimator, the reverse process can, in principle, approximate any target action distribution to arbitrary precision. Recent analysis [37] shows that, when the score-matching and discretization errors are bounded, the KL-divergence between the learned diffusion policy and the target multimodal distribution can be driven below any ε. Effectively, the diffusion process behaves like an infinitely-expressive Gaussian-mixture, embedding an unrestricted number of modes without the need to pre-specify mixture components.

For CD&R this multimodal capacity is crucial. Confronted with the same conflict scenario, a unimodal policy may only "learn to turn right"; a diffusion policy can represent several safe maneuver classes simultaneously—turn right, hold course, descend, climb—each corresponding to a probability peak. At execution time the agent retains viable options: if right-hand airspace is blocked (weather, restricted area) it can still select a left-hand or vertical maneuver with high probability, thereby increasing safety-of-resolution and overall robustness.

### 4.1.2 Value-Guided Score Matching

Pure maximum-entropy training can scatter probability mass over unsafe or low-value actions, which may lead to inefficient exploration. To steer the multimodal distribution toward high-return regions, we adopt a Q-value-weighted score-matching objective. In this formulation, the denoising network is trained not only to match the true score but also to amplify gradients in proportion to the estimated state–action value $Q(s,a)$. This mechanism functions as a value-guided policy gradient under a maximum-entropy regularizer, encouraging the network to preserve action diversity while directing probability mass towards high-reward peaks.

Compared to hand-crafted Gaussian-mixture policies, our approach avoids mode-collapse and the need for priors on the number of modes. The modal structure emerges directly from the environment's dynamics and reward signal, as the Q-values naturally guide the distribution towards high-value regions. In this way, the teacher distribution used in the policy update (in §4.3) is inherently shaped by the environment's dynamics and the task's value structure, without any need for manually specified mixture components.

### 4.1.3 Unified Treatment of Factorized Discrete Action Spaces

A second motivation for adopting diffusion policies is their ability to model the joint distribution over factorized discrete actions. In our task, the action is a 3D intention triple ($\Delta\psi,\Delta v,\Delta h$) with each component in {-1,0,+1}, yielding 27 classes. The diffusion denoising network directly parameterizes a joint categorical distribution over these 27 options, capturing cross-dimension dependencies without ad-hoc heads or hierarchies. Unlike unimodal Gaussian policies, this joint modeling typically induces a multimodal action distribution, concentrating probability mass on several coordinated maneuver modes and thereby improving diversity.

#### 4.1.4 Safety and Interpretability Considerations

Performance alone is insufficient in safety-critical ATM applications; policies must be auditable and verifiably safe. First, the stochastic diffusion policy can be distilled into an interpretable surrogate—decision trees or symbolic rules—providing post-hoc explanations for controller validation. Second, safety is enforced during training via (i) large penalties on loss-of-separation (LoS) and near-mid-air-collision (NMAC) events, (ii) conservative dual-Q critics that down-weight risky maneuvers, and (iii) a density-progressive safety curriculum (Section 4.4) that exposes the agent to increasingly dense traffic only after it masters low-density scenarios. Together, these mechanisms align the expressive power of diffusion policies with a safety-conscious, interpretable RL paradigm.

In summary, our diffusion policy framework directly tackles the three challenges from Section 1. It achieves superior action diversity through multimodal policies capable of generating rich maneuver sets. This is enabled by a novel, unified head that parameterizes the joint distribution over the factorized discrete space of 27 actions, preserving vital cross-dimension coupling. Furthermore, we ensure safety and transparency with value-guided training, curriculum learning, and post-hoc distillation. The following sections will formalize the Diffusion Actor-Critic architecture and Density-Progressive Safety Curriculum that operationalize these advantages.

### 4.2 Diffusion Policy Architecture

This section details the diffusion-policy network that maps a state $s$ to a joint distribution over 27 discrete action tuples $a = (a^\psi, a^v, a^h) \in \{-1, 0, 1\}^3$. The three intention components correspond to heading change, speed change, and flight-level step, respectively; their physical execution uses fixed per-step increments defined in Section 3 (e.g. $\Delta\psi_{step} = 5°$, $\Delta v_{step} = 50kt$; see Eqs.(16)-(20)). To achieve expressive, multimodal decision-making while preserving cross-dimension coupling, we parameterize the policy as a conditional denoising diffusion model in 27-dimensional logit space. Concretely, the network predicts "clean" logits $y_0 \in \mathbb{R}^{27}$ whose softmax yields a joint categorical distribution $\pi_\theta(\cdot \mid s) = softmax(y_0(s))$. Diffusion is used to learn $y_0(s)$ by inverting a Gaussian noising process on logits; this provides the capacity to place probability mass on several coordinated maneuver modes(e.g., a left turn with $a+1$ flight-level step and, alternatively, an acceleration with level hold), overcoming the unimodal bias of conventional policies.

#### 4.2.1 State conditioning and network backbone

At each decision epoch, the agent observes s, which includes ownship kinematics, relative states of the three nearest intruders, waypoint/route context, and auxiliary indicators defined in §3.3.1. A state encoder Enc(s) (a small MLP with layer-norm and Mish nonlinearity) produces an embedding h=Enc(s). We adopt a UNet-style denoising backbone with residual blocks and self-attention, closely following Fig. 4: a symmetric encoder–decoder with skip connections, down / up sampling stages (e.g., channel widths 128→256→512 in the encoder and 512→256→128 in the decoder), and 8-head self-attention at the bottleneck to let the model focus on conflict-critical features. A sinusoidal time-step embedding $\gamma(t)$ conditions every residual block so that the network behaves appropriately across the $T$ denoising steps (coarse edits at large $t$, fine refinements near $t = 0$). We set the diffusion depth to a moderate $T = 10$ steps for real-time control with sufficient expressiveness, and use a linear noise schedule $\{\beta_t\}_{t=1}^{T}$ that increases the logit-space variance gradually; these choices match our training procedure in §4.3 and the runtime constraints of the CD&R simulator.

**4.2.2 Joint-categorical parameterization in logit space**

We represent the policy over the 27 joint action classes—each class encoding an intention tuple $(a^{\psi}, a^{v}, a^{h}) \in \{-1, 0, +1\}^3$ by clean logits $y_0(s) \in \mathbb{R}^{27}$ whose softmax gives a joint categorical distribution. Diffusion is carried out in logit space (rather than on one-hot vectors or probabilities) so that training and inference share the same tensorial interface and numerical stability is preserved. Formally, we endow the 27-D logit vector $y_t$ with a DDPM-style forward (noising) process $q_\beta$ controlled by a variance schedule $\{\beta_t\}_{t=1}^{T}$ and define the associated reverse (denoising) process $p_\theta$ conditioned on the state $s$.

We first specify the forward kernel that corrupts logits one step at a time. With $\alpha_t = 1 - \beta_t \in (0,1)$ and the identity $I \in \mathbb{R}^{27 \times 27}$, the forward transition and its marginal take the standard Gaussian form:

$$q_\beta(y_t | y_{t-1}) = \mathcal{N}\left(\sqrt{\alpha_t} y_{t-1}, (1-\alpha_t)I\right) \tag{27}$$

$$q_\beta(y_t | y_0) = \mathcal{N}\left(\sqrt{\bar{\alpha}_t} y_0, (1-\bar{\alpha}_t)I\right) \tag{28}$$

where $\bar{\alpha}_t = \prod_{i=1}^{t} \alpha_i$ accumulates the schedule multiplicatively. We also use the customary DDPM quantity

$$\tilde{\beta}_t = \frac{1 - \bar{\alpha}_{t-1}}{1 - \bar{\alpha}_t} \beta_t \tag{29}$$

with $\tilde{\beta}_t$ appearing in the covariance of the reverse kernel.

The reverse dynamics in logit space are parameterized by a denoising network $D_\theta$ that predicts the

clean logits from a noised input at step $t$. Writing

$$\hat{y}_0 = D_\theta(s, y_t, t) \tag{30}$$

we define a Gaussian reverse kernel

$$p_\theta(y_{t-1} | y_t, s) = \mathcal{N}\left(\mu_\theta(y_t, s, t), \tilde{\beta}_t I\right) \tag{31}$$

whose mean is the standard $x_0$-prediction combination

$$\mu_\theta(y_t, s, t) = \frac{\sqrt{\bar{\alpha}_{t-1}} \beta_t}{1 - \bar{\alpha}_t} \hat{y}_0 + \frac{\sqrt{\alpha_t}(1 - \bar{\alpha}_{t-1})}{1 - \bar{\alpha}_t} y_t \tag{32}$$

Although each reverse transition (31) is Gaussian, the mean $\mu_\theta(y_t, s, t)$ is a state- and time-conditioned nonlinear function learned by the denoiser; chaining these transitions therefore pushes the base Gaussian $y_T \sim \mathcal{N}(0, I)$ to a generally non-Gaussian (often multimodal) distribution over $y_0$.

The policy is obtained by applying a safety-masked softmax to the predicted clean logits. Let $\mathcal{A}_{\text{sfe}}(s) \subseteq \{1, \ldots, 27\}$ denote the set of action classes that satisfy aircraft-envelope and immutable procedural rules under state $s$. This feasibility set is state-local and outcome-agnostic, i.e., it is independent of inter-aircraft separation geometry; we then normalize only over feasible classes:

$$\pi_\theta(a | s) = \frac{\exp(\hat{y}_0(a)) \mathbf{I}[a \in \mathcal{A}_{\text{sfe}}(s)]}{\sum_{a'} \exp(\hat{y}_0(a')) \mathbf{I}[a' \in \mathcal{A}_{\text{sfe}}(s)]}, a \in \{1, \ldots, 27\} \tag{33}$$

We do not mask actions merely because they may reduce separation; LoS/NMAC risk is handled by termination and reward shaping (§3, §4.4). Speed rate limiting is applied at state update rather than by masking.

At inference, we initialize $y_T \sim \mathcal{N}(0, I)$, iterate $y_{t-1} \sim p_\theta(y_{t-1} | y_t, s)$ for $t = T, \ldots, 1$, apply the masked softmax (33) to obtain $\pi_\theta(\cdot | s)$, and select a class by argmax sampling; the chosen class is then decoded to $(a^\psi, a^v, a^h)$ and executed with the fixed step sizes defined in §3. At training time, the same interface is used, with $y_0$ supplied by a value-guided teacher distribution (see §4.3), ensuring that §4.2 and §4.3 form a mathematically consistent pipeline.

**4.2.3 Class decoding and physical execution**

Given the masked joint distribution $\pi_\theta(\cdot | s)$ in §4.2.2, we obtain a class $\hat{a} \in \{1, \cdots, 27\}$ by argmax sampling. Each class indexes a unique intention triple $(a^\psi, a^v, a^h) \in \{-1, 0, +1\}^3$ via a fixed lookup table aligned with the factorization order $(\Delta\psi, \Delta v, \Delta h)$. The triple is mapped to per-step physical increments using constants defined in §3, after which the kinematics are updated with rate limiting on speed.

This decoding closes the loop from logits to executable maneuvers and keeps §4.2 numerically consistent with the MDP and the fixed step sizes specified in §3, while the safety mask $\mathcal{A}_{\text{sfe}}(s)$ in §4.2.2 guarantees envelope-compliant choices prior to execution.

**4.2.4 Training–inference interface**

The denoising backbone in §4.2.2 is fed with the same-shaped tensors in training and inference; only the source of $y_t$ differs. At training time, a value-guided, safety-masked teacher distribution $p^*(a|s)$ (constructed from the critics in §4.3) is transformed into clean logits $y_0(s) = \ln(p^*(\cdot|s) + \varepsilon/27) \in \mathbb{R}^{27}$ and then corrupted to $y_t$ by the forward process $q_\beta$. The denoiser predicts $\hat{y}_0 = D_\theta(s, y_t, t)$ and is optimized with a logit-space denoising loss plus an optional terminal cross-entropy against $p^*$. At inference time, the same network starts from Gaussian logit noise $y_T \sim \mathcal{N}(0, I)$, iterates the learned reverse kernel $p_\theta(y_{t-1}|y_t, s)$ for $T$ steps to obtain $y_0$, applies the safety-masked softmax to form $\pi_\theta(\cdot|s)$, selects a class, and executes it as in §4.2.3. This concise contract makes §4.2–4.3 mathematically contiguous and prevents ambiguity about what is fed to or produced by the model in each mode. The architecture of proposed diffusion policy network is shown in **Fig. 4**.

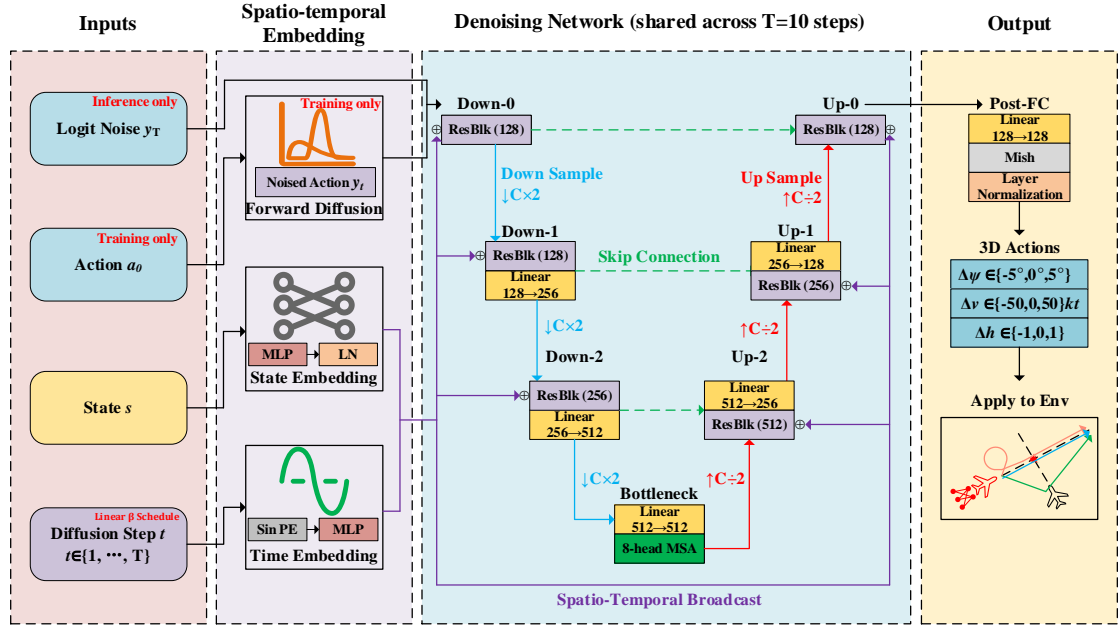

**Fig. 4 Architecture of Diffusion Policy Network**

## 4.3 Training Procedure for Diffusion Policy and Critics

We train the diffusion policy and critics in an off-policy actor–critic fashion with experience replay. The policy is the masked joint categorical defined in §4.2: for state $s$, the denoiser $D_\theta$ produces clean logits $\hat{y}_0(s) \in \mathbb{R}^{27}$; after the safety mask $\mathcal{A}_{\text{sfe}}(s)$ is applied, the policy is $\pi_\theta(a|s) \propto \exp(\hat{y}_0(a)) \mathbf{I}[a \in \mathcal{A}_{\text{sfe}}(s)]$. The critics approximate the state–action value on the discrete action set $\{-1,0,+1\}_{\Delta\psi} \times \{-1,0,+1\}_{\Delta v} \times \{-1,0,+1\}_{\Delta h}$ (27 classes). We maintain two critics $Q_{\phi_1}, Q_{\phi_2}$ with slowly updated targets $Q_{\bar{\phi}_1}, Q_{\bar{\phi}_2}$ (Polyak averaging) to avoid over-estimation and to stabilize bootstrapping. Throughout, $\gamma \in (0,1)$ denotes the discount, $d \in \{0,1\}$ is the terminal indicator, $r$ is the reward, and $\mathcal{D}$ is the replay buffer that stores tuples $(s, a, r, s', d)$.

### 4.3.1 Critic learning on a masked 27-class action set

Given a mini-batch $\{(s_i, a_i, r_i, s'_i, d_i)\}_{i=1}^{N}$ sampled uniformly from $\mathcal{D}$, we compute a target for each transition using the masked maximization backup on the discrete next-state actions. Let $\mathcal{A}_{\text{sfe}}(s') \subseteq \{1, \ldots, 27\}$ be the feasible classes at $s'$. We define

$$y_i = r_i + \gamma(1 - d_i) \max_{a' \in \mathcal{A}_{\text{sfe}}(s'_i)} \min\left\{Q_{\bar{\phi}_1}(s'_i, a'), Q_{\bar{\phi}_2}(s'_i, a')\right\} \tag{34}$$

where the inner $\min\{\cdot,\cdot\}$ is the double-Q conservative target and the outer max is computed exhaustively over the 27 masked actions (no approximation is needed because $|\mathcal{A}| = 27$). This is computationally tractable and avoids the variance associated with Monte Carlo action sampling for target evaluation.

The critics minimize the squared temporal-difference (TD) error

$$\mathcal{L}_Q(\phi_1, \phi_2) = \frac{1}{N} \sum_{i=1}^{N} \left[ \left(Q_{\phi_1}(s_i, a_i) - y_i\right)^2 + \left(Q_{\phi_2}(s_i, a_i) - y_i\right)^2 \right] \tag{35}$$

After each update we softly update the targets by Polyak averaging,

$$\bar{\phi}_j \leftarrow \tau_{\text{polyak}} \phi_j + (1 - \tau_{\text{polyak}}) \bar{\phi}_j, \; j \in \{1,2\} \tag{36}$$

with $\tau_{\text{polyak}} \in (0,1)$. Equation (34)-(36) define a stable Bellman update tailored to the discrete masked action set, fully consistent with the bounded-dimension logits in §4.2.2. The double-Q target together with Polyak-averaged target networks mitigates over-estimation and stabilizes the discrete masked backup in (34)-(36).

### 4.3.2 Value-guided teacher and diffusion policy objective

To align the policy with high-value yet diverse maneuvers, we construct for each $s$ a soft teacher distribution over the 27 classes from the current critics, and train the denoiser to reconstruct its clean logits under diffusion. Let $Q_\phi(s,a) = \min\{Q_{\phi_1}(s,a), Q_{\phi_2}(s,a)\}$. With a temperature $\tau > 0$ that controls sharpness and the safety mask $\mathcal{A}_{\text{sfe}}(s)$, we define

$$p^*(a|s) = \frac{\exp(Q_\phi(s,a)/\tau)\mathbf{1}[a \in \mathcal{A}_{\text{sfe}}(s)]}{\sum_{a'}\exp(Q_\phi(s,a')/\tau)\mathbf{1}[a' \in \mathcal{A}_{\text{sfe}}(s)]}, a \in \{1,\cdots,27\} \quad (37)$$

where the teacher temperature $\tau > 0$ controls sharpness: smaller $\tau$ concentrates mass on top-valued classes, whereas larger $\tau$ preserves multimodality across several high-value maneuver modes. In integrated training loop, $\tau$ annealed with curriculum stage.

We transform this teacher into clean logits $y_0(s)$ in the 27-D logit space introduced in §4.2 by a numerically stable log-map with small smoothing $\varepsilon > 0$:

$$y_0(s) = \ln(p(\cdot|s) + \varepsilon/27) \in \mathbb{R}^{27} \quad (38)$$

where the small smoothing constant $\varepsilon$ prevents ln0 and improves numerical stability of the logit-space diffusion objective.

For diffusion training we sample a step $t \sim \mathcal{U}\{1,\ldots,T\}$, corrupt $y_0$ to $y_t$ by the forward kernel $q_\beta$ defined in §4.2.2, and ask the denoiser to predict $\hat{y}_0 = D_\theta(s, y_t, t)$. The policy loss combines a logit-space denoising regression with an optional terminal cross-entropy that matches the teacher at $t=0$:

$$\mathcal{L}_\pi(\theta) = \frac{1}{N}\sum_{i=1}^{N}\mathbb{E}_{\substack{t \sim \mathcal{U}\{1,\ldots,T\} \\ \varepsilon \sim \mathcal{N}(0,I)}}[\|D_\theta(s_i, y_{t,i}, t) - y_{0,i}\|_2^2] + \lambda_{\text{CE}}\frac{1}{N}\sum_{i=1}^{N}\text{CE}(p^*(\cdot|s_i), \text{softmax}(D_\theta(s_i, y_{0,i}, 0))) \quad (39)$$

where:

$$y_{t,i} = \sqrt{\bar{\alpha}_t}\, y_{0,i} + \sqrt{1-\bar{\alpha}_t}\,\varepsilon, \quad (40)$$

$$y_{0,i} = \log(p^*(\cdot|s_i) + \varepsilon_{\text{sm}}/27). \quad (41)$$

Here $y_{t,i}$ is the noised logit for sample $i$ generated via the forward kernel $q_\beta(y_t|y_0)$ in §4.2.2, $\text{CE}(\cdot|\cdot)$ is cross-entropy on the 27-class simplex, and $\lambda_{\text{CE}} > 0$ sets the strength of the terminal classifier consistency. The loss (39) preserves multimodality by regressing toward a soft teacher (37) rather than a one-hot label; the temperature $\tau$ controls the entropy of $p^*$, and the safety mask prevents probability mass on envelope-violating classes. We use the same diffusion depth $T$ and noise schedule $\{\beta_t\}$ as in §4.2, so so training and inference operate with shape-identical tensors and comparable per-step computational cost.

**4.3.3 End-to end Diffusion–AC training flow**

Each training iteration interleaves critic and policy updates in a way that mirrors the generation chain defined in §4.2. In practice, we draw a mini-batch from $\mathcal{D}$ and update critics by minimizing (35) with targets (34); we then hold the critic parameters fixed, for the same batch of states, compute the value-guided teacher (37), form clean logits via (38), sample diffusion steps and noised logits using the forward process $q_\beta$, and finally update the denoiser by minimizing (39). Afterward we softly update $Q_{\bar{\phi}_1}, Q_{\bar{\phi}_2}$ using (36). This procedure constitutes a single Diffusion–Actor–Critic step and is repeated at a fixed ratio to environment interaction (see §4.4). In all phases the safety mask $\mathcal{A}_{\text{sfe}}(\cdot)$ is applied consistently: in targets (34) to avoid bootstrapping on infeasible actions, in the teacher (37) to avoid supervising illegal classes, and in the policy's masked softmax in §4.2.2 to avoid selecting them at inference. The result is an online algorithm that shifts probability mass toward high-value, safe modes without sacrificing the multi-peak structure that is essential for robust CD&R. Algorithm 1 shows a typical training iteration.

---

**Algorithm 1** Diffusion–Actor–Critic (DAC): One Training Iteration

---

**Call-time contract.** *By construction*, the interface specification, diffusion/guidance hyperparameters, and optimizer/target-update settings **[G2]–[G4]** are fixed as global constants at initialization and remain unchanged throughout training. Consequently, invoking this algorithm for a single iteration requires only **[G1]**.

**Require: [G1 — Call-time core]** Mini-batch $\{(s_i, a_i, r_i, s'_i, d_i)\}_{i=1}^{N}$; learnable params $Q_{\phi_1}, Q_{\phi_2}$ with targets $\bar{\phi}_1, \bar{\phi}_2$; denoiser $D_\theta$; discount $\gamma \in (0, 1)$.

**Require: [G2 — Interface spec]** Discrete joint head of size $K=27$; upstream feasibility mask $m_{\mathrm{sfe}}(s) \in \{0,1\}^K$, masked set $\mathcal{A}_{\mathrm{sfe}}(s) = \{k : m_k = 1\}$, and masked softmax $\mathrm{softmax}_{\mathcal{A}}(z)_k = \dfrac{e^{z_k} \mathbb{1}[k \in \mathcal{A}]}{\sum_j e^{z_j} \mathbb{1}[j \in \mathcal{A}]}$.

**Require: [G3 — Diffusion & guidance hyperparams]** Linear schedule $\{\beta_t\}_{t=1}^T$ with $\alpha_t = 1 - \beta_t$, $\bar{\alpha}_t = \prod_{u=1}^{t} \alpha_u$; temperature $\tau > 0$; smoothing $\varepsilon_{\mathrm{sm}} > 0$; CE weight $\lambda_{\mathrm{CE}} \ge 0$.

**Require: [G4 — Optimizer & target-update]** Stepsizes $\eta_Q, \eta_\pi$; Polyak coefficient $\rho \in (0, 1)$.

**Ensure:** Updated $\phi_1, \phi_2, \bar{\phi}_1, \bar{\phi}_2, \theta$ after one iteration; training–inference interface parity holds.

    **(A) Critic update: conservative backup over feasible actions**

1: *Rationale:* compute a masked, double-$Q$ target so that value guidance later reflects only physically/legally feasible classes.
2: **for** $i = 1$ to $N$ **do**
3:     $\mathcal{A}' \leftarrow \mathcal{A}_{\mathrm{sfe}}(s'_i)$
4:     $q_i \leftarrow \max\limits_{k \in \mathcal{A}'} \min \{ Q_{\bar{\phi}_1}(s'_i, k), Q_{\bar{\phi}_2}(s'_i, k) \}$
5:     $y_i \leftarrow r_i + \gamma(1 - d_i) q_i$
6: **end for**
7: $\mathcal{L}_Q \leftarrow \dfrac{1}{N} \sum_{i=1}^{N} \left[ (Q_{\phi_1}(s_i, a_i) - y_i)^2 + (Q_{\phi_2}(s_i, a_i) - y_i)^2 \right]$
8: $\phi_1 \leftarrow \phi_1 - \eta_Q \nabla_{\phi_1} \mathcal{L}_Q$;    $\phi_2 \leftarrow \phi_2 - \eta_Q \nabla_{\phi_2} \mathcal{L}_Q$

    **(B) Value-guided teacher in probability space**

9: *Rationale:* turn $Q$ into a multimodal teacher $p^\star(\cdot \mid s)$ confined to the feasible set and controlled by temperature $\tau$.
10: Keep critics fixed and define $Q_\phi(s, k) = \min\{Q_{\phi_1}(s, k), Q_{\phi_2}(s, k)\}$.
11: **for** $i = 1$ to $N$ **do**
12:     $\mathcal{A} \leftarrow \mathcal{A}_{\mathrm{sfe}}(s_i)$
13:     $p_i^\star(k) \leftarrow \dfrac{\exp(Q_\phi(s_i, k)/\tau) \mathbb{1}[k \in \mathcal{A}]}{\sum_j \exp(Q_\phi(s_i, j)/\tau) \mathbb{1}[j \in \mathcal{A}]}$
    *Label-smoothed clean logits in* logit *space*
14:     $y_{0,i} \leftarrow \ln(p_i^\star + \varepsilon_{\mathrm{sm}}/K) \in \mathbb{R}^K$     ▷ avoids $\ln 0$ and stabilizes tails
15: **end for**

    **(C) Logit-space forward diffusion (training-time only)**

16: *Rationale:* corrupt clean logits to a random time $t$ so the denoiser learns to predict $y_0$ from $y_t$.
17: **for** $i = 1$ to $N$ **do**
18:     sample $t \sim \mathcal{U}\{1, \ldots, T\}$ and $\varepsilon \sim \mathcal{N}(0, I_K)$
19:     $y_{t,i} \leftarrow \sqrt{\bar{\alpha}_t} \, y_{0,i} + \sqrt{1 - \bar{\alpha}_t} \, \varepsilon$
20: **end for**

    **(D) Denoiser/policy update with terminal masked likelihood**

21: *Rationale:* combine denoising regression at time $t$ with a terminal cross-entropy that matches the teacher on the masked head.
22: $\mathcal{L}_{\mathrm{diff}} \leftarrow \dfrac{1}{N} \sum_{i=1}^{N} \| D_\theta(s_i, y_{t,i}, t) - y_{0,i} \|_2^2$
23: $\mathcal{L}_{\mathrm{term}} \leftarrow \dfrac{1}{N} \sum_{i=1}^{N} \mathrm{CE}\left( p_i^\star, \mathrm{softmax}_{\mathcal{A}_{\mathrm{sfe}}(s_i)}\left( D_\theta(s_i, y_{0,i}, 0) \right) \right)$
24: $\mathcal{L}_\pi \leftarrow \mathcal{L}_{\mathrm{diff}} + \lambda_{\mathrm{CE}} \mathcal{L}_{\mathrm{term}}$
25: $\theta \leftarrow \theta - \eta_\pi \nabla_\theta \mathcal{L}_\pi$

    **(E) Target critics: Polyak averaging**

26: **for** $j \in \{1, 2\}$ **do**
27:     $\bar{\phi}_j \leftarrow \rho \, \phi_j + (1 - \rho) \, \bar{\phi}_j$
28: **end for**

    **Notes for reproducibility (inference parity).**

29: *At inference*, start from $y_T \sim \mathcal{N}(0, I_K)$ and apply the same denoiser schedule $T \to 0$; decode with the *same* masked head $\mathrm{softmax}_{\mathcal{A}_{\mathrm{sfe}}(s)}(\cdot)$, ensuring training–inference interface equivalence.

---

## 4.4 Integrated Agent-Environment Interaction Loop

This section integrates the Diffusion–AC policy with the simulator under a density-progressive safety curriculum. The action interface remains identical to §§4.2–4.3: decisions are sampled from the masked 27-class joint categorical obtained from logit-space denoising, and feasibility is enforced upstream by the stage-invariant mask $\mathcal{A}_{\text{sfe}}(s)$. The curriculum changes only the scenario density and the LoS/NMAC termination and reward shaping defined in §3; the policy head, mask, and update rules are unchanged.

**4.4.1 Curriculum objective and safety contract**

We employ a density-progressive safety curriculum that increases encounter difficulty while keeping the policy/safety interface fixed. The action space and parameterization are exactly those of §4.2: the policy runs in 27-D logit space, produces clean logits $y_0(s) \in \mathbb{R}^{27}$, and forms a masked joint categorical $\pi_\theta(\cdot | s)$ by normalizing only over the feasible classes $\mathcal{A}_{\text{sfe}}(s)$. This feasibility mask encodes hard aircraft-envelope and rule constraints and is applied upstream—before normalization and selection—so that infeasible classes never receive probability mass; no post-hoc action correction is used. The curriculum modifies only the scenario density supplied by the traffic generator, the termination thresholds used to detect Loss-of-Separation (LoS) and Near-Mid-Air Collision (NMAC), and the relative weights assigned to the reward components already defined in §3. Crucially, the mask $\mathcal{A}_{\text{sfe}}(s)$ itself is stage-invariant across the curriculum. With this contract, the masked maximization used by the critics (§4.3.1), the value-guided teacher (§4.3.2), and the masked softmax policy (§4.2.2) remain interface-consistent at every stage.

**4.4.2 Schedules for density and safety shaping**

We run a twelve-stage curriculum with index $k \in \{1, \ldots, 12\}$. Stage advancement is gated by a fixed window of $W = 100$ episodes: let $\mathfrak{S}(e) \in \{0,1\}$ indicate a successful deconfliction episode under stage $k$. The rolling success rate over the last $W$ episodes at stage $k$ is:

$$\hat{p}_k(t) = \frac{1}{W} \sum_{e=t-W+1}^{t} \mathfrak{S}(e). \tag{42}$$

We promote from stage $k$ to $k+1$ if $\hat{p}_k(t) \geq 0.90$; otherwise training continues at stage $k$.

Safety shaping tightens only through the LoS termination threshold, while the NMAC standard remains constant as defined in §3. The LoS distance increases linearly from 4.5km at stage $k = 1$ to 10km at $k = 12$:

$$d_{\text{LoS}}(k) = 4.5 + 0.5(k-1) \text{km} \tag{43}$$

$$d_{\text{NMAC}}(k) = 0.2 \text{km} \tag{44}$$

These thresholds affect episode termination and reward shaping only; they do not modify the feasibility mask used in action selection.

Traffic density increases in lock-step with the stage. The scenario generator instantiates $R(k)$ active routes and a total of $A(k)$ intruder aircraft according to the linear schedule

$$R(k) = k, \quad A(k) = 3k, \tag{45}$$

which moves from $(R,A) = (1,3)$ at $k = 1$ to $(12,36)$ at $k = 12$. This density schedule alters the state distribution faced by the agent but leaves the policy head, the feasibility mask $\mathcal{A}_{sfe}(s)$, and the logit-space denoising interface unchanged from §4.2.2.

### 4.4.3 Agent–environment interaction loop during training

At curriculum stage k (defined by the density and safety schedules in §4.4.2), each episode proceeds as an MDP with tuples $(s_t, a_t, r_t, s_{t+1}, d_t)$. At time t, the agent observes s_t and selects an action through the fixed logit-space inference interface of §4.2.2: we draw $y_T \sim \mathcal{N}(0, I_{27})$, run the learned reverse denoising process conditioned on $s_t$ to obtain clean logits $y_0$, apply the stage-invariant feasibility mask $\mathcal{A}_{sfe}(s_t)$ and normalize to form the masked joint categorical $\pi_\theta(\cdot \mid s_t)$, and choose the class by argmax. The selected class is decoded to $(a^\psi, a^v, a^h) \in \{-1, 0, +1\}^3$ and executed with the fixed step sizes of §3; the kinematics update deterministically with speed rate-limiting only as specified there, while heading and flight level follow the fixed increments. The environment computes $r_t$ using the stage-k reward weights and applies the stage-k LoS/NMAC termination criteria; it then produces $s_{t+1}$ and the terminal flag $d_t$. This loop repeats until $d_t=1$ or the episode horizon is reached.

Transitions are appended to the replay buffer and drive off-policy learning. At a fixed update ratio (as in §4.3), mini-batches from replay are used to update the critics and the diffusion policy without altering the interaction protocol: the masked exhaustive maximization over the 27 classes defines conservative double-Q targets, the critics minimize the squared TD loss with Polyak target updates, and the policy is trained against the value-guided teacher via the logit-space denoising objective. In all three places—target construction, teacher formation, and policy normalization—the same feasibility mask $\mathcal{A}_{sfe}(\cdot)$ is applied upstream, so the training and acting interfaces are shape-identical in $\mathbb{R}^{27}$. Stage advancement follows the performance gate from §4.4.2 and affects only the scenario density and the LoS termination threshold for subsequent episodes; the 27-class policy head, the denoising schedule, and the masking contract remain unchanged. Algorithm 2 is pseudocode of the agent-environment interaction during training loop.

---

**Algorithm 2** Agent–Environment Interaction under DPSC (Training Loop)
---
**Require:** Policy denoiser $D_\theta$; critics $Q_{\phi_1}, Q_{\phi_2}$ and targets $\bar{\phi}_1, \bar{\phi}_2$; replay buffer $\mathcal{D}$; curriculum stages $k \in \{1,\ldots,12\}$ with density/safety schedules $R(k), A(k), d_{\text{LoS}}(k)$ and constant $d_{\text{NMAC}}$; window $W=100$ and promotion threshold $p_{\text{adv}}=0.90$; update period $U$ (e.g., $U=4$); fixed inference interface of § 4.2.2 and rewards/terminations of § 3.
**Ensure:** Trained $\theta, \phi_1, \phi_2, \bar{\phi}_1, \bar{\phi}_2$; policy consistent with § 4.2–§ 4.4.
1: *Stage k configured by § 4.4.2*
2: **for** $k = 1$ **to** 12 **do**
3:    Configure scenario generator with $R(k)$ routes and $A(k)$ intruders; set thresholds $d_{\text{LoS}}(k)$ and $d_{\text{NMAC}}$.
4:    *episodes observed at stage k*
5:    $n_k \leftarrow 0$
6:    **while not** PROMOTED **do**
7:      Reset environment; receive initial state $s_0$; $t \leftarrow 0, d_0 \leftarrow 0$
8:      **while** $d_t = 0$ **and** $t < T_{\max}$ **do**
9:        *(Acting; fixed logit-space interface of § 4.2.2)*
10:        Sample $y_T \sim \mathcal{N}(0, I_{27})$; run reverse denoising conditioned on $s_t$ to obtain clean logits $y_0$
11:        Form masked policy $\pi_\theta(\cdot \mid s_t) \propto \mathbf{1}[a \in \mathcal{A}_{\text{sfe}}(s_t)] \exp(y_0[a])$; set $a_t \leftarrow \arg\max_a \pi_\theta(a \mid s_t)$
12:        Decode $a_t \mapsto (\Delta\psi, \Delta v, \Delta h) \in \{-1, 0, +1\}^3$; apply fixed step sizes of § 3 and speed rate-limiting (Eq. (20)) to update kinematics
13:        Environment returns $(r_t, s_{t+1}, d_t)$ using stage-$k$ reward weights and termination tests with $d_{\text{LoS}}(k)$, $d_{\text{NMAC}}$
14:        Push transition $(s_t, a_t, r_t, s_{t+1}, d_t)$ into replay buffer $\mathcal{D}$;   $s_t \leftarrow s_{t+1}$
15:        **if** $(t \mod U) = 0$ **then**
16:           *Learning; single Diffusion–AC update by Algorithm 1*
17:           $(\theta, \phi_1, \phi_2, \bar{\phi}_1, \bar{\phi}_2) \leftarrow$ ALGORITHM 1$(\mathcal{D}; \theta, \phi_1, \phi_2, \bar{\phi}_1, \bar{\phi}_2)$
18:        **end if**
19:        $t \leftarrow t + 1$
20:      **end while**
21:      $n_k \leftarrow n_k + 1$;   $\mathfrak{S}_{k,n_k} \leftarrow \mathbf{1}\{$episode completed without LoS/NMAC and task criteria of § 3 satisfied$\}$
22:      **if** $n_k \geq W$ **then**
23:        *rolling success over last $W=100$ episodes at stage k*
24:        $\widehat{p}_k \leftarrow \frac{1}{W} \sum_{e=n_k-W+1}^{n_k} \mathfrak{S}_{k,e}$
25:        **if** $\widehat{p}_k \geq p_{\text{adv}}$ **then**
26:           PROMOTED $\leftarrow$ true
27:        **end if**
28:      **end if**
29:    **end while**
30: **end for**

---

# 5 Case Study

## 5.1 Experiment Design and Parameters Setting

### 5.1.1 3D Airspace Configuration

To simulate typical conflict scenarios within an area control airspace, we constructed a simulation environment encompassing a three-dimensional space and multiple airways. Horizontally, the airspace is defined as a square area of 2000x2000 units, corresponding to approximately 400 km x 400 km in the real world. For ease of numerical computation and visualization, real-world distances are normalized using a scale factor where one simulation unit approximates 200m. Vertically, the airspace is discretized into three flight levels (FL0, FL1, and FL2), with a standard separation of 300m between adjacent levels. Within this 3D space, a network of fixed airways is pre-defined to serve as flight paths.

These routes are formed by connecting pairs of points on the airspace boundaries and are designed to intersect, emulating the structure of a busy airway network. For our concurrent traffic simulation, we selected 12 representative, straight-line airways that traverse the airspace and cross other routes, thereby constituting a high-density, high-complexity airway structure.

Building upon this static airspace model, we designed a dynamic traffic scenario to generate a persistent stream of intruder aircraft. The environment procedurally spawns an intruder on a fixed airway at the start of each of the 12 selected routes. Initially, one intruder cruises at a constant speed along each airway at the lowest flight level, FL0. As the simulation progresses, a phased generation strategy is employed based on route completion. When an intruder at FL0 covers approximately one-third of its trajectory, a second intruder is generated at the same entry point at FL1. Subsequently, when the FL0 intruder reaches the two-thirds progress mark, a third intruder is spawned at FL2. This phased generation strategy ensures that up to three aircraft can be simultaneously active on each route at different flight levels, following one another at a set distance and uniform speed. This process creates dense, multi-layered traffic flows across the entire airspace. Upon completing its route, each intruder exits the airspace boundary and is removed. A new intruder is then immediately respawned at FL0 on the same route to maintain the continuity of the traffic flow. Through this cyclic mechanism, the simulation consistently sustains a high-density traffic state of approximately 36 aircraft (12 airways × up to 3 layers), providing a complex environment to train the agent's autonomous conflict detection and resolution capabilities.

At the beginning of each training episode, the agent-controlled aircraft is randomly initialized at a point on the airspace boundary and assigned a random waypoint (Goal). These waypoints are uniformly distributed within the interior of the airspace, at least 20 km from any boundary to avoid trivial tasks, representing the final destination of a mission assigned by air traffic control. As the agent navigates from its initial position toward this goal, it must execute avoidance maneuvers in response to the continuously generated stream of intruder aircraft. To prevent immediate conflicts at the start of an episode, the environment guarantees an initial separation of at least 20 km between the agent and the nearest intruder.

The kinematics of both the agent and intruder aircraft are simulated in discrete time steps, where each step $\Delta t$ corresponds to approximately 30 seconds of real-world time. In each step, the positions of all aircraft are updated based on their current velocity and heading. Intruder aircraft maintain a constant cruise speed of 800 km/h with a fixed heading along their designated airways. In contrast, the agent can adjust its own heading and speed at every time step. The airspace boundaries are treated as hard constraints; if the agent crosses these boundaries, it is considered a rule violation and incurs a penalty.

**5.1.2 Simulation Scenario Parameters**

To quantify the key parameters of the simulation scenario, we define the following settings for traffic flow and safety separation standards:

(1) Number of Intruders and Perception Limits: The environment maintains 12 initial intruder aircraft (one for each of the 12 airways) and dynamically replenishes them using the mechanism described previously, resulting in an average of approximately 36 intruders present for the agent to avoid in each episode. The agent's observation space, however, is limited to information from only the three nearest intruders. This constraint simulates the limited situational awareness of a human air traffic controller or an onboard system, reducing the state dimensionality while focusing the agent's attention on the most imminent threats.

(2) Generation Frequency: The spawning of new intruders is triggered by route progress thresholds. The time interval between two consecutive intruders on the same airway is determined by their cruise speed and the route length, with the average frequency set such that a new aircraft appears when the preceding one has completed one-third of its journey.

(3) Initial Separation: As previously mentioned, the agent's initial position is guaranteed to be no less than 100 units (approximately 20 km) from any intruder. The initial separation between two consecutive intruders on the same route is also maintained at this scale. This ensures that no immediate, close-proximity conflicts occur at the start of an episode or at the moment of spawning, providing the agent with adequate reaction time.

(4) Loss of Separation (LoS) Standard: In line with operational standards for area control, the horizontal separation threshold, $d_{\text{LoS}}$, is set to 10 km. A Loss of Separation (LoS) event is declared if the agent and an intruder are at the same flight level (or if the agent is transitioning and the intruder is at its original or target level) and their horizontal distance is less than 10 km. When an LoS event occurs, the episode does not terminate immediately; instead, the conflict is recorded, and the agent receives a penalty.

(5) Near Mid-Air Collision (NMAC) Criterion: Based on CCAR-93TM-R6, the NMAC criterion, $d_{\text{NMAC}}$, is defined as a horizontal distance of 0.2 km. An NMAC is considered to have occurred if two aircraft are at the same flight level and their separation is less than this distance. This is treated as a critical conflict, analogous to a physical collision, and results in the immediate termination of the episode. Thus, 10 km and 0.2 km serve as the minimum safety separation and collision thresholds in this simulated environment, respectively.

(6) Time Step Limit: The maximum number of simulation steps per training episode is set to 1000. If the agent fails to reach its goal within this limit without an NMAC, the episode is terminated due to a timeout.

Collectively, these parameter settings create a simulation environment that maximizes the frequency and complexity of potential conflicts while preserving a necessary safety margin, thereby generating rich and diverse decision-making scenarios for subsequent algorithm training.

**5.1.3 Algorithm Training Parameters**

To address the demands of the aforementioned airspace environment and task, we designed and implemented a deep reinforcement learning algorithm based on a diffusion model policy. The training procedure and model hyperparameters are configured as follows. The training protocol consists of 10,000 simulation episodes, with each episode running for a maximum of 1000 environment steps; episodes exceeding this limit are terminated due to a timeout.

The algorithm's exploratory behavior is fundamentally driven by the intrinsic stochasticity of its policy network. In the initial phase of training, the diffusion model generates actions from pure Gaussian noise, guaranteeing broad exploration of the action space. As training progresses, the policy gradually converges under the joint influence of Q-function guidance. However, it crucially retains its capacity to represent a multimodal action distribution, thereby striking a sophisticated balance between Exploration and Exploitation.

The experience replay buffer is configured with a capacity of $10^6$ transitions. We use a mini-batch size of 128 and perform a parameter update every four environment steps (a training frequency of 4). Each update consists of a single gradient descent step, meaning a batch of 128 transitions is sampled for one learning step every four interactions with the environment. We employ the Adam optimizer, with a learning rate of $3\times10^{-4}$ for the diffusion policy network and $10^{-3}$ for the critic networks. The discount factor ($\gamma$) is set to 0.99. To ensure training stability, we maintain target networks for both the policy and the two critic networks, which are updated via a soft target update with a coefficient ($\tau$) of $5\times10^{-3}$. To ensure the reproducibility of our study, **Table 1** summarizes the five core hyperparameter configurations most critical to performance. The selection of these values is experimentally justified in the sensitivity analysis detailed in Section 5.3.2.

**Table 1 Key Hyperparameters in Experiments**

| Parameter Category | Parameter Name | Symbol | Value |
|---|---|---|---|
| Diffusion Policy | Diffusion Steps | $T$ | 10 |
| Value and Policy Update | Q-Value Weighting Temperature | $w_{temp}$ | 1 |
|  | Soft Update Coefficient | $\tau$ | $5\times10^{-3}$ |
| NN Architecture | Critic Network Capacity | - | 128 |

**5.1.4 Diffusion Policy Model Configuration**

The agent's policy network is architected as a diffusion probabilistic model, which iteratively generates collision avoidance maneuvers by denoising from a Gaussian prior. The diffusion process is discretized into 10 diffusion steps (T=10). The noise scale schedule, $\beta_t$, follows a linear scheme, increasing uniformly from $\beta_0 = 1\times10^{-4}$ to $\beta_T = 2\times10^{-2}$. At each diffusion step $t$, the policy network takes the current state and a time step embedding as input to predict and remove noise from the action representation. The network itself is a feed-forward neural network that integrates residual connections and multi-head self-attention. First, the time step t undergoes sinusoidal positional encoding and is passed through a multi-layer perceptron (MLP) to extract a time embedding vector. This vector is then concatenated with the state vector (processed by a state encoder to produce a hidden representation) and the current action noise. The combined tensor is processed through a sequence of four fully-connected layers featuring LayerNorm and Mish activation functions to progressively extract features. To enhance feature propagation, this backbone is augmented with four residual blocks. Furthermore, an 8-head self-attention mechanism is incorporated into the hidden layers to capture dependencies between the different dimensions of the action space. For the value-estimation component, we employ a dual-critic architecture to separately evaluate the goal-achievement value and the safety-risk value of the agent's actions. Each critic is a three-layer, fully-connected neural network that takes the state vector as input and outputs the corresponding Q-values for each action. The hidden layer dimensionality of the critics is set to 128. In line with the twin-critic strategy, the minimum of the two Q-values is used to guide the policy update, yielding a conservative evaluation of safety while still optimizing for returns. Through the synergy of this diffusion policy network and the dual-critic architecture, the agent can efficiently learn effective conflict resolution strategies within the high-dimensional, hybrid action space.

**5.1.5 Reward Function Settings**

Grounded in the MDP formulation from Section 3, we designed a reward function that explicitly balances mission objectives with safety requirements, comprising both goal-oriented rewards and safety-related penalties. A terminal reward of +1.0 is granted upon successful, collision-free arrival at the goal waypoint, incentivizing the agent to complete its mission safely. To promote efficiency and discourage protracted or inefficient maneuvers, a minor per-step penalty of -0.001 is applied at each time step. Conversely, the reward structure incorporates several penalties for unsafe or unsuccessful behavior. A significant penalty of -0.5 is incurred for each Loss of Separation (LoS) event, defined as a breach of the 10 km safety standard. While this penalizes the violation, the episode continues, allowing the agent the opportunity to recover. In stark contrast, a critical conflict of NMAC severity (<0.2 km) is treated as a catastrophic collision. This results in the immediate termination of the episode and imposes a substantial penalty of -1.0. Furthermore, episode failure due to an airspace boundary violation or a timeout (exceeding 1000 steps) also incurs a negative reward, typically equivalent to the penalty for a single LoS event. This comprehensive reward mechanism guides the agent to learn a policy that intrinsically weighs the trade-off between safety and efficiency. The positive reward for success is counterbalanced by penalties for inefficiency and, most critically, for safety violations, compelling the agent to find an optimal strategy that is both effective and robust.。

**5.1.6 Evaluation Index Design**

To comprehensively evaluate the performance of our proposed algorithm on the complex airspace conflict resolution task, we adopt the following metrics, which are aligned with the methodology in Section 3 and the experiments in Section 4:

1) Goal Achievement Rate: Defined as the proportion of episodes in which the agent successfully reaches its assigned waypoint without a collision. This metric, equivalent to the mission success rate, directly reflects the algorithm's ability to fulfill the primary control objective: guiding an aircraft safely to its destination. A higher goal achievement rate indicates that the agent can effectively resolve conflicts and complete its mission under a wide range of circumstances.

2) Loss of Separation (LoS) Rate: The proportion of episodes in which at least one LoS event—a breach of the 10 km separation standard—occurs between the agent and any intruder aircraft. This metric quantifies the frequency of safety violations during the resolution process. Ideally, the LoS rate should be as low as possible. It is important to note that this rate encompasses all episodes with an LoS event, including those that ultimately end in success as well as those that time out, thereby providing a measure of the policy's overall safety margin.

3) Near Mid-Air Collision (NMAC) Rate: The proportion of episodes terminated due to an NMAC event, where the distance between the agent and an intruder falls below the critical 0.2 km threshold. This metric is a direct measure of the algorithm's ability to prevent catastrophic failures and can be considered a key component of the overall failure rate. A lower NMAC rate signifies a safer and more reliable policy.

4) Timeout Rate: The proportion of episodes that conclude because the maximum step limit is reached without the agent having either achieved its goal or experienced an NMAC. This metric serves as an indicator of policy effectiveness; a high timeout rate often implies that the agent has learned an inefficient strategy involving excessive or unproductive maneuvers and is thus also considered a form of mission failure.

5) Average Step Count: The average number of steps taken by the agent to successfully complete its mission, calculated over all goal-achieved episodes. This metric reflects the efficiency and maneuver overhead of the learned policy. A lower average step count suggests a more efficient strategy, where the agent resolves conflicts with minimal deviation. Conversely, a higher value may indicate longer detours or less timely maneuvers.

By evaluating these metrics collectively, we can conduct a holistic assessment of the agent's performance across three critical dimensions: Success (Goal Achievement Rate), Safety (LoS and NMAC Rates), and Efficiency (Timeout Rate and Average Step Count). In the subsequent analysis, these metrics will form the basis for a quantitative comparison and discussion of our proposed algorithm against the selected baselines.

**5.1.7 Computer Software and Hardware Resources**

The training and testing for this study were conducted on a high-performance computing platform. The primary hardware components include an Intel Core i9-13900K CPU, an NVIDIA GeForce RTX 4090 GPU with 24 GB of VRAM, and 128 GB of system memory. This configuration provides ample computational power and storage for the large-scale training of deep reinforcement learning algorithms, ensuring that the intensive matrix operations and parallel processing required by the diffusion model can be executed efficiently with GPU acceleration. The software environment was built on Python 3.10 and the PyTorch 2.0.0 deep learning framework, in conjunction with the CUDA 11.8 toolkit to leverage GPU acceleration during the training process.

**5.2 Benchmark Performance Evaluation**

To conduct a comprehensive and objective evaluation of our proposed conflict resolution algorithm based on a diffusion probabilistic model (hereafter referred to as Diffusion-AC), we selected several representative and high-performing algorithms from the deep reinforcement learning field to serve as baselines. These algorithms were benchmarked in an identical simulation environment, as detailed in Section 5.1. The primary objective of this comparative analysis is to validate whether our proposed method offers significant advantages over conventional DRL frameworks—particularly in terms of success rate, safety, and convergence efficiency—when addressing multi-agent conflict problems in a three-dimensional, hybrid action space.

We selected the following four archetypal algorithms as our baselines:

1) Safe-DQN-X [9]: An extension of DQN that employs a dual-network architecture, with one network prioritizing safety constraints and the other focusing on goal achievement, designed to enhance policy safety and interpretability.

2) Rainbow-DQN [7]: An advanced DQN variant that integrates multiple improvements, including Double DQN, Dueling Networks, Prioritized Experience Replay, and multi-step learning, enabling more efficient learning of discrete avoidance policies.

3) Proximal Policy Optimization (PPO) [49]: A leading on-policy policy gradient algorithm renowned for its stable training performance, which it achieves by clipping the policy update to ensure monotonic improvement. For this benchmark, we customized a policy network for PPO capable of handling a hybrid action space (continuous heading/speed and discrete altitude) by using separate output heads for continuous and discrete action components.

4) Twin-Delayed Deep Deterministic Policy Gradient (TD3) [50]: A state-of-the-art off-policy actor-critic algorithm that mitigates Q-value overestimation through the use of clipped double Q-learning and delayed policy updates. As TD3 natively supports continuous actions, we adapted it to our hybrid action space by mapping the discrete altitude component to an additional continuous output dimension, which was then discretized to the nearest flight level during execution.

All deep reinforcement learning algorithms (Diffusion-AC, PPO, TD3, Safe-DQN-X, and

Rainbow-DQN) were trained under identical experimental conditions: 10,000 training episodes, an experience replay buffer capacity of 1,000,000, and a unified reward function, state space, and set of environment parameters. To ensure a fair comparison, we meticulously tuned the key hyperparameters for each algorithm (e.g., learning rates, network architectures) to achieve their optimal performance. The final evaluation of each algorithm was conducted after its policy had converged, based on statistical results from an additional 1,000 inference episodes to mitigate the effects of stochasticity. The implementations of the baseline algorithms were based on open-source codebases or pseudocode provided in their respective papers: PPO was implemented using the Stable-Baselines3 library, while TD3, Safe-DQN-X, and Rainbow-DQN were based on their official open-source implementations. For fairness, all algorithms employed a similar three-layer neural network architecture (128 units per layer) and the same discount factor of 0.99. Specific parameters, such as PPO's clip ratio of 0.2 and GAE coefficient of 0.95, and TD3's actor/critic learning rates and target update coefficient, were set to their recommended values after tuning for optimal stability.

**5.2.1 Comparison of Training Convergence Process**

**Fig. 5** presents the raw learning curves for the Combined Reward of the five algorithms over the course of 10,000 training episodes. As illustrated, all algorithms exhibit a rapid initial ascent within the first 500 episodes, climbing from a very low starting reward (approx. -1500) to a significantly higher performance level (approx. -150 to -100). Among them, PPO (orange) demonstrates the steepest learning curve, approaching its performance plateau around episode 180 and consistently maintaining the highest reward (close to -80) thereafter. Diffusion-AC (blue) follows closely; although its initial reward peak occurs slightly later than PPO's (around episode 220), it stabilizes at approximately -110 by episode 2,000 and displays a smoother, more gradual upward trend in the subsequent training. It is worth noting that this phenomenon of rapid initial convergence does not contradict the progressive philosophy of DPSC. The low-density scenarios in the early stages of the curriculum naturally yield higher rewards, allowing both Diffusion-AC and PPO to achieve high scores quickly. The later phases of training, which introduce progressively higher traffic densities, correspond to the slow fine-tuning stage observed in both curves.

In contrast, Rainbow-DQN (green) and Safe-DQN-X (red) exhibit a markedly gentler learning slope. Rainbow-DQN only begins to approach the performance plateau of Diffusion-AC after 5,000 episodes, while Safe-DQN-X experiences a pronounced oscillation around the 4,000-episode mark before entering a stable phase. TD3 (purple) stagnated near a reward of -800 throughout the training process, exhibiting high-frequency variance, which is indicative of severely inefficient exploration within the hybrid action space.

In terms of final performance at 10,000 episodes, the ranking of Combined Reward is as follows: PPO ≈ Safe-DQN-X ≈ Diffusion-AC ≈ Rainbow ≫ TD3. Although PPO achieves a slightly higher final reward, Diffusion-AC demonstrates a more significant advantage in key safety metrics such as NMAC and LoS rates, as will be detailed in the following section. In summary, while PPO shows the strongest convergence capability in terms of raw reward, Diffusion-AC rapidly approaches a stable performance plateau early in training and maintains a low-variance, steady state thereafter, laying a robust foundation for the subsequent evaluation of its safety and efficiency.

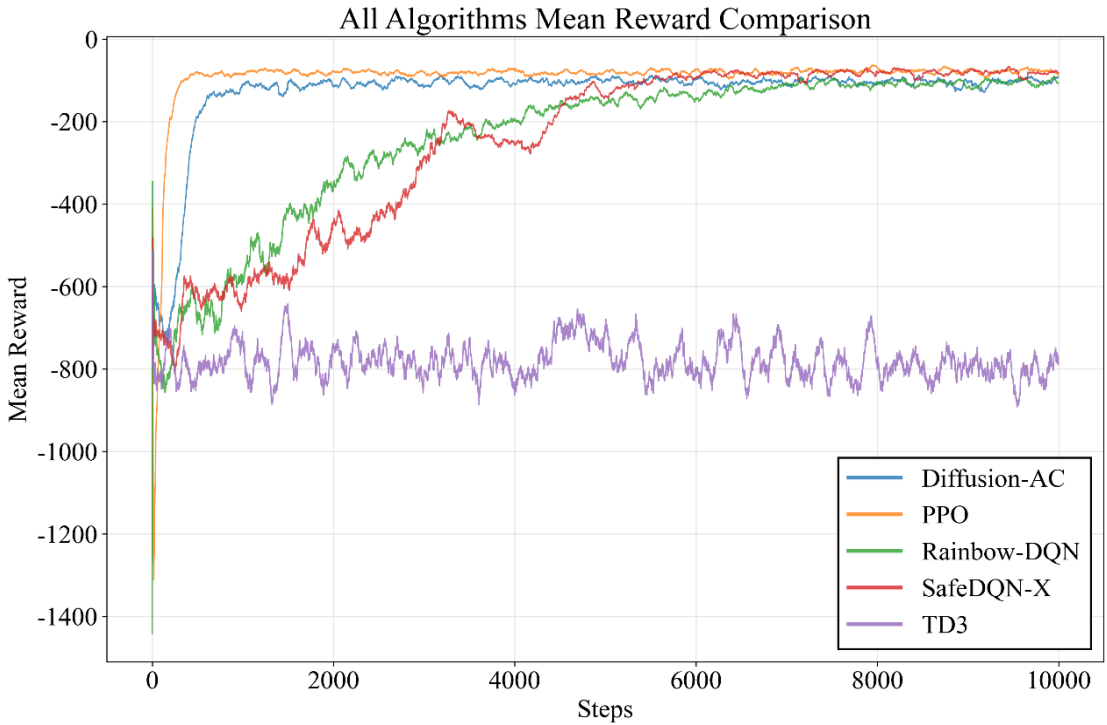

Fig. 5 Training Convergence Comparison

5.2.2 Inference Performance Comparison

During the inference phase, we evaluated the comprehensive performance of each algorithm under varying traffic densities. Based on the environment settings from Section 5.1, we adjusted the intruder generation frequency to create test scenarios with 24 (low-density), 36 (medium-density), and 48 (high-density) intruder aircraft. Each density level was evaluated over 1,000 independent episodes, during which we recorded the goal achievement rate, LoS rate, NMAC rate, timeout rate, average step count, and average reward. The results are summarized in **Table 2**.

**Table 2 Inference Performance Comparison**

| Algorithm | Traffic Density | Success Rate (%) | LoS Rate (%) | NMAC Rate (%) | Timeout Rate (%) | Average Steps | Average Goal Reward | Average Safe Reward |
|---|---|---|---|---|---|---|---|---|
| Diffusion-AC | Low | **100** | **0.2** | **0** | **0** | 258 | **-96.44** | **-0.15** |
|  | Medium | **96.9** | **0.7** | **0.3** | 2.8 | 279 | **-96.83** | **-1.04** |
|  | High | **94.1** | **2.8** | **1.2** | 4.7 | 302 | **-95.8** | **-1.98** |
| PPO | Low | **100** | 0.7 | **0** | **0** | 251 | -98.06 | -0.53 |
|  | Medium | 96 | 8 | 3.9 | **0.1** | **272** | -100.48 | -1.32 |
|  | High | 93.3 | 10.7 | 6.4 | **0.3** | 310 | -98.05 | -2.14 |
| TD3 | Low | 5.6 | 18.6 | 7.1 | 87.3 | 910 | -705.01 | -2.57 |
|  | Medium | 4.4 | 23.4 | 4.6 | 91 | 948 | -732.83 | -2.22 |
|  | High | 4.3 | 27.1 | 8 | 87.7 | 932 | -696.69 | -3.57 |
| Safe-DQN-X | Low | **100** | 0.9 | **0** | **0** | 310 | -100.7 | -0.68 |
|  | Medium | 93.8 | 4.3 | 2 | 4.2 | 369 | -99.61 | -1.84 |
|  | High | 91.2 | 7.2 | 3 | 5.8 | 383 | -98.72 | -2.87 |
| Rainbow-DQN | Low | **100** | 0.8 | **0** | **0** | **244** | -110.43 | -0.61 |
|  | Medium | 92.5 | 7.4 | 4 | 3.3 | 343 | -108.81 | -2.2 |
|  | High | 90.3 | 11.3 | 5.4 | 4.3 | 368 | -110.37 | -2.72 |

The inference results demonstrate that Diffusion-AC consistently maintains the optimal balance between safety and efficiency across all three traffic densities. In the low-density scenario, while Diffusion-AC, PPO, Safe-DQN-X, and Rainbow-DQN all achieved a 100% success rate, Diffusion-AC suppressed the LoS rate to a mere 0.2% with a 0% NMAC rate. By comparison, the next best performer, PPO, still recorded a 0.7% LoS rate, with Safe-DQN-X and Rainbow-DQN at 0.9% and 0.8%, respectively. As traffic density increased to medium, Diffusion-AC's success rate only marginally decreased to 96.9%, with LoS and NMAC rates rising to just 0.7% and 0.3%. In stark contrast, PPO's LoS rate surged to 8%, with its NMAC rate reaching 3.9%, while the LoS rates for Safe-DQN-X and Rainbow-DQN remained in the 4-7% range. The high-density scenario most clearly delineates the performance gap: Diffusion-AC maintained a 94.1% success rate, an LoS rate of 2.8%, and an NMAC rate of 1.2%. Meanwhile, PPO's LoS and NMAC rates escalated to 10.7% and 6.4%, respectively, while Rainbow-DQN's rates climbed as high as 11.3% and 5.4%. Across all densities, TD3 failed to produce a viable policy, with a success rate below 6% and the vast majority of episodes ending in timeouts or collisions, confirming its inability to handle the hybrid action space.

The efficiency metrics further substantiate the superiority of Diffusion-AC. As traffic density increased from low to high, its average step count grew by only a small margin, from 258 to 302 steps, indicating its resolution strategies remain highly efficient even in congested airspace. By comparison,

PPO required 310 steps in the high-density scenario, while Safe-DQN-X and Rainbow-DQN needed 383 and 368 steps, respectively. TD3's average step count consistently exceeded 900, with over 85% of its episodes timing out, demonstrating a near-total loss of resolution capability. In terms of reward, Diffusion-AC's average goal reward in the high-density scenario remained at -95.8, significantly outperforming PPO (-98.05) and Safe-DQN-X (-98.72). Concurrently, the absolute value of the safety penalty incurred by the Diffusion-AC policy was only 1.98, far lower than the 2.1–2.9 range of the other algorithms.

In conclusion, Diffusion-AC achieves the optimal combination of success, safety, and performance across all metrics. It ensures a near-100% mission success rate while simultaneously minimizing the risks of mid-air conflicts and collisions and maintaining high trajectory efficiency. The underlying reason for this superior performance is its ability to maintain a multimodal policy distribution. When faced with sudden changes in the conflict geometry or maneuver constraints, it can rapidly switch to second- or third-best avoidance actions, thus preventing the policy rigidity that leads to collisions in other models. Conversely, the policies learned by PPO and the DQN family of algorithms tend to collapse into a unimodal or fragmented mode; once their primary strategy fails, they lack viable alternatives, causing a sharp increase in safety violations at high densities. Furthermore, Diffusion-AC integrates safety-aware mechanisms such as Q-weighted denoising and a density-progressive curriculum during training, which strictly prune high-risk actions and gradually increase exploratory difficulty. These design choices enable a more effective balance between exploration and exploitation during policy optimization. Consequently, even when confronted with complex environments, Diffusion-AC maintains robust decision-making, comprehensively outperforming traditional baseline algorithms in both safety and operational efficiency.

**5.2.3 Decision Delay and Real-time Analysis**

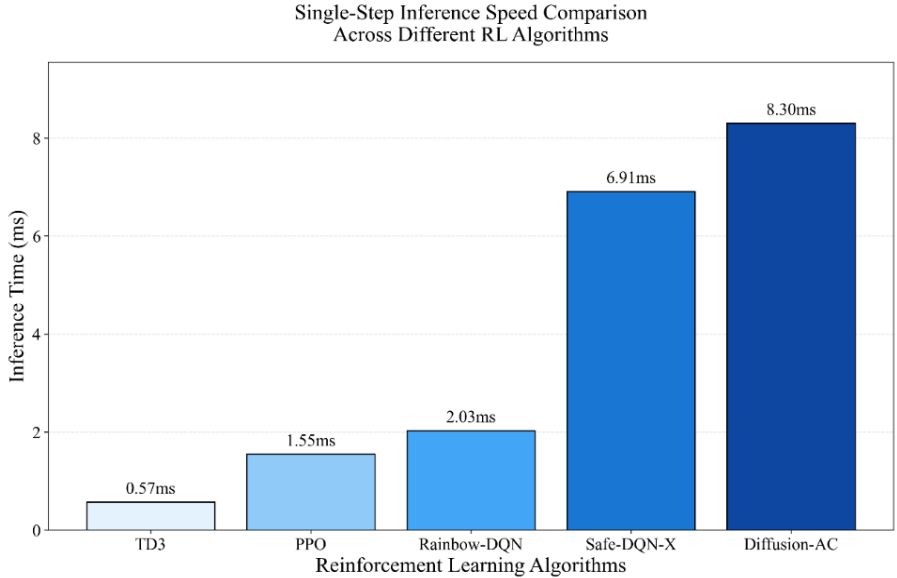

**Fig. 6 Single-Step Inference Speed Comparison Across Different RL Algorithms**

In addition to policy effectiveness, we also compared the inference speed of each algorithm. In practical air traffic control applications, the computational latency of a single decision step is a critical factor that directly impacts the real-time viability and deployability of an algorithm. We measured the average single-step inference latency for each algorithm on the same hardware platform (Intel i9-13900K CPU, RTX 4090 GPU).

The results reveal significant differences in decision-making time. Conventional algorithms based on feed-forward neural networks, such as PPO, TD3, and Rainbow-DQN, all exhibit latencies in the range of 0.6 to 2 milliseconds. Rainbow-DQN is slightly slower (~2 ms) due to its more complex architecture (incorporating dueling and noisy layers), while the simpler models of PPO and TD3 are faster, at approximately 1.5 ms and 0.6 ms, respectively. Safe-DQN-X, which maintains an additional safety evaluation network, has a slightly higher inference overhead, increasing its latency to around 7 ms.

In comparison, Diffusion-AC's single-step decision latency is approximately 8.3 ms, which is markedly higher than the baseline algorithms. This is primarily due to the iterative sampling process required by the diffusion policy at each decision point. In our implementation, the reverse denoising process involves T=10 steps, meaning that each action selection requires 10 forward passes through the U-Net architecture. Even so, a decision latency of under 10 ms is still well within acceptable limits for ATC applications, where the feedback loop for conflict detection and resolution is typically on the order of seconds. This ensures that Diffusion-AC can provide timely resolution commands. However, the higher computational cost does mean that its current inference efficiency is lower than that of more lightweight algorithms. In a practical deployment, this latency could be further reduced through techniques such as model distillation, network pruning, or the use of dedicated hardware accelerators.

Overall, the bar chart in **Fig. 6** clearly illustrates the average single-step decision times for each algorithm. Although Diffusion-AC has a slightly higher latency, it remains within a practical operational envelope, whereas baselines like PPO and TD3 are exceptionally fast. The trade-off between Diffusion-AC's inference speed and its superior performance, along with potential optimization pathways, will be discussed in subsequent sections.

## 5.3 Ablation Study

### 5.3.1 Key Components Ablation Analysis

To comprehensively evaluate our proposed Diffusion-AC framework, we conducted a series of exhaustive ablation studies. These studies were designed not only to validate the effectiveness of specific components we introduced, such as the DPSC, but also to examine the necessity and impact of the core mechanisms derived from Diffusion-RL—namely, value guidance and entropy regularization—within the context of the air traffic conflict resolution problem. We systematically removed or replaced several core modules of the Diffusion-AC framework, resulting in the following five "W/O (without)" model variants:

1. W/O Diffusion Policy: In this variant, the diffusion-based policy is replaced with a standard Gaussian policy network. Consequently, the agent directly outputs actions from a unimodal Gaussian distribution instead of generating them through an iterative diffusion process. This ablation is designed to isolate and evaluate the contribution of the diffusion policy architecture itself to the overall performance.

2. W/O Dual Q: The dual-Q critic architecture in Diffusion-AC is replaced with a single Q-network. This ablation serves to verify the role of the dual-critic design in mitigating value overestimation and enhancing training stability.

3. W/O DPSC: We remove the Density-Progressive Safety Curriculum. Instead of progressively increasing traffic density and penalty severity, the agent is trained from the outset using the most challenging, high-density scenarios and the full set of penalties. This setup evaluates the impact of curriculum learning on convergence stability and final performance.

4. W/O Value Guidance: This variant assesses the guiding role of the value signal in the diffusion

process. We achieve this by removing the Q-value weighting term from the actor's loss function , effectively fixing the temperature τ to infinity. As a result, the learning of the diffusion policy is no longer directly driven by the critic's value estimates. This ablation examines the importance of value guidance for accelerating convergence and focusing exploration on high-value action regions.

**Table 3** summarizes the performance of each ablation variant in the high-density test scenario (48 intruder aircraft), benchmarked against the complete Diffusion-AC algorithm. The evaluation metrics include success rate, NMAC rate, timeout rate, and average step count.

Table 3 Key Components Ablation Analysis

| Algorithm Variants | Success Rate (%) | LoS Rate (%) | NMAC Rate (%) | Timeout Rate (%) | Average Steps |
|---|---|---|---|---|---|
| Diffusion-AC (Complete) | 94.1 | 2.8 | 1.2 | 4.7 | 302 |
| W/O Diffusion | 89.2 | 12.3 | 6.4 | 4.4 | 342 |
| W/O Double Q | 91.3 | 4.5 | 2.1 | 6.6 | 323 |
| W/O DPSC | 87.6 | 6.4 | 2.9 | 9.5 | 350 |
| W/O Value Guidance | 4.1 | 26.2 | 11.2 | 84.7 | 953 |

The results in **Table 3** clearly show that removing the diffusion-based policy significantly undermines strategy diversity and directly magnifies conflict risk. Without the diffusion module (W/O Diffusion), the policy degenerates to a unimodal Gaussian output, causing the success rate to drop to 89.2%. Concurrently, the LoS and NMAC rates climb sharply to 12.3% and 6.4%, respectively, and the average step count increases to 342. This demonstrates that the inherent multimodal expressiveness provided by the diffusion architecture is crucial for maintaining behavioral redundancy; its absence leads to a contraction of the action distribution, a reduction in viable alternative maneuvers, and consequently, a substantial increase in the probability of collisions in high-density environments.

In contrast, removing the dual-Q critic (W/O Double-Q) primarily impacts the conservatism of value estimation. A single-critic model is more susceptible to optimistic value estimates for high-risk actions, leading to an increase in the NMAC rate from 1.2% to 2.1% and the LoS rate to 4.5%, with a corresponding 3-percentage-point drop in the success rate. Although the average step count increases only slightly (from 302 to 323), the safety margin is clearly diminished, confirming the critical role of the dual-Q architecture in suppressing overestimation and promoting safe decision-making.

Ablating the density-progressive curriculum (W/O DPSC) forces the agent to confront high-density traffic and strict penalties from the very beginning of training. This drastically increases the difficulty of exploration and slows convergence. The final model achieves a success rate of only 87.6%, with the NMAC rate rising to 2.9% and the timeout rate climbing to 9.5%, while the average step count increases to 350. These results indicate that a curriculum-based, progressive-difficulty approach not only improves sample efficiency but also significantly reduces collision risk in the final high-density stages. Its absence causes the policy to become trapped in a suboptimal local minimum.

The most severe degradation in performance occurs with the removal of value guidance (W/O Value Guidance). Without Q-weighting, the diffusion sampling process is unable to distinguish between high- and low-value actions, causing the policy to become almost entirely random. The success rate plummets to a mere 4.1%, while the NMAC rate skyrockets to 11.2% and the timeout rate reaches 84.7%, with the average step count approaching 1000. This catastrophic failure unequivocally demonstrates that value guidance is the central mechanism of reinforcement learning with diffusion models; its removal renders the learning process completely ineffective.

In summary, the superior performance of Diffusion-AC is a synergistic result of the behavioral redundancy afforded by its multimodal diffusion policy, the safety-oriented value landscape constructed through conservative dual-Q evaluation and Q-weighting, and the structured exploration provided by the DPSC curriculum. This integrated design ensures that the algorithm can simultaneously achieve a high success rate and a minimal rate of critical conflicts, even in extremely high-density airspace.

**5.3.2 Key Hyperparameters Sensitivity Ablation Analysis**

To conduct an in-depth investigation into the influence of various hyperparameters on the performance of the Diffusion-AC algorithm, we systematically adjusted the number of diffusion steps (T), the Q-value weighting temperature ($w_{\text{temp}}$), the soft update coefficient ($\tau$), and the critic network capacity. For each hyperparameter, we performed five independent training and inference runs for each of its configured values, recording the corresponding changes in key performance metrics. As shown in **Fig. 7**, the shaded areas represent the variance across these five experimental runs.

Firstly, as shown in **Fig. 7(a)**, our analysis of the number of diffusion steps (T) reveals that the model fails to learn an effective policy when T<8. Increasing T from 8 to 10 results in a substantial jump in the conflict resolution success rate, from 84.62% to 94.15%, while the single-step inference latency increases from 6.81 ms to 8.32 ms. Further increases to T=20, 40, and 80 yield marginal gains in success rate (to 94.87%, 95.46%, and 95.85%, respectively) but at the cost of a dramatic, near-exponential increase in inference latency (to 18.44 ms, 37.84 ms, and 89.41 ms). This indicates that the improvement in policy quality plateaus for T⩾10, while the computational overhead grows prohibitively. Therefore, a balanced choice of T=10 ensures a high success rate while maintaining acceptable real-time decision-making capability.

As shown in **Fig. 7(b)**, For the Q-value weighting temperature ($w_{\text{temp}}$), increasing the temperature from 0.1 to 1.0 boosts the success rate from 90.92% to 94.10%, though the NMAC rate also rises from 0.88% to 1.20%. A further increase to 5.0 results in a marginal improvement in success rate (to 94.28%) but a continued increase in the NMAC rate (to 1.57%). An excessively high temperature of 10.0 causes performance to degrade, with the success rate falling to 92.39% and the NMAC rate climbing to 2.42%. This trend suggests that a moderate temperature in the range of 1.0 to 5.0 strikes an effective balance between value guidance and policy diversity; temperatures that are too low or too high are detrimental to performance.

The sensitivity analysis of the soft update coefficient ($\tau$) in **Fig. 7(c)** shows that a very small $\tau$ ($10^{-3}$) yields a stable average reward of -100.4 with low variance (std. dev. 6.41). Increasing $\tau$ to $5 \times 10^{-3}$ improves the average reward to -95.8 but also increases the variance (std. dev. 8.46). The peak average reward of -95.38 is achieved at $\tau=10^{-2}$, with a further increase in variance (std. dev. 10.61). An overly large $\tau$ (0.05) leads to a drop in average reward to -96.37 and a sharp increase in variance (std. dev. 16.92). These results indicate that a moderate update rate (between $5 \times 10^{-3}$ and $10^{-2}$) balances learning stability with reward maximization, whereas target networks that update too slowly or too quickly can negatively impact training.

Finally, as shown in **Fig. 7(d)**, the critic network capacity significantly affects value function approximation. With only 64 hidden units, the success rate is 91.67% and the NMAC rate is 1.82%. Increasing the capacity to 128 units boosts the success rate to 94.11% and reduces the NMAC rate to 1.22%. The highest success rate of 94.31% is achieved with 256 units (NMAC rate ~1.40%). A further increase to 512 units results in a slight performance drop. Thus, a critic network with a hidden layer size between 128 and 256 possesses sufficient capacity for accurate value approximation while maintaining training stability and avoiding the high-variance issues associated with oversized models.

In summary, this hyperparameter sensitivity analysis reveals that a diffusion step count of T=10, a Q-value temperature between 1.0 and 5.0, a soft update coefficient in the range of $5 \times 10^{-3}$ to $10^{-2}$, and a critic hidden dimension of 128–256 provide an optimal configuration. This combination achieves an excellent balance among conflict resolution success rate, NMAC rate, and model runtime efficiency.

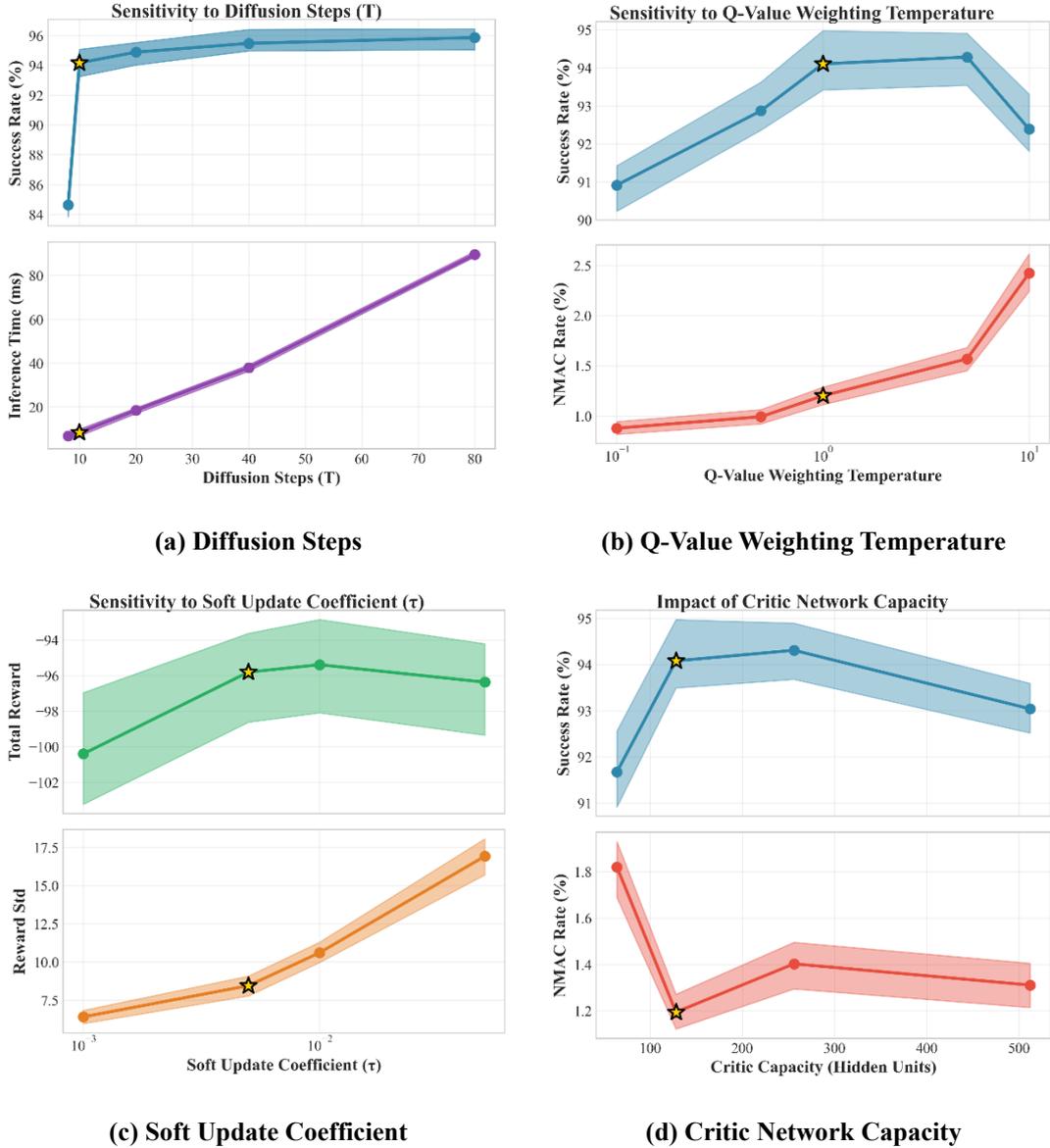

(a) Diffusion Steps  (b) Q-Value Weighting Temperature

(c) Soft Update Coefficient  (d) Critic Network Capacity

Fig. 7 Key Hyperparameters Sensitivity Ablation Analysis

## 5.4 Multimodal Action Distribution Validation

This section aims to quantitatively and intuitively validate the core advantage of the Diffusion-AC model through visualization: its ability to generate multimodal and diverse resolution commands in a three-dimensional action space. A common limitation of traditional deep reinforcement learning algorithms, such as PPO, is that their action distributions tend to "collapse" to a single optimal solution during policy optimization, rendering them rigid in complex and dynamic real-world environments. The central hypothesis of our research is that Diffusion-AC can overcome this deficiency by retaining and outputting multiple logically sound and safe resolution strategies for the same conflict scenario.

To focus the analysis on the model's core capabilities in complex decision-making environments, we will exclude simple one-on-one conflicts that can be easily resolved with a single vertical maneuver. The solution space for such scenarios is relatively fixed and too simplistic to reveal the deeper differences in policy diversity and decision-making robustness among various reinforcement learning algorithms. Therefore, this experiment is centered exclusively on a series of complex tactical scenarios with meticulously crafted constraints. These scenarios are designed to test and showcase the model's ability to generate fine-grained composite maneuvers, handle asymmetric constraints, and effectively trade off between multiple viable solutions. By conducting a direct visual comparison with a typical unimodal policy model, we aim to intuitively demonstrate the significant advantage of Diffusion-AC in terms of decision-making flexibility.

**5.4.1 Experiment Settings**

To systematically analyze the model's decision-making behavior and ensure the rigor of our conclusions, our experimental design employs a two-stage framework: (1) Policy Distribution Sampling and (2) Policy Efficacy Validation.

In the first stage, we place the agent at a precise "critical decision point"—a state where, if no action is taken, a conflict is imminent within the next 60 to 90 seconds. From this critical state, we perform large-scale Monte Carlo inference, recording only the first action generated by the model in each run. The purpose of this step is to capture the model's most primitive and complete decision-making tendencies in a pure and unbiased manner when facing an initial, high-stakes conflict.

However, the generated action distribution alone is insufficient; its effectiveness must be validated. Therefore, in the second stage, each unique "first-step action" sampled in the first stage is subjected to a full follow-on simulation. In these simulations, the agent executes the initial action and then continues to operate until the conflict is either successfully resolved or the episode ends in failure. A trial is labeled as a "failure" and terminated immediately upon the occurrence of an LoS, an NMAC, or a timeout. Through this process, we assign a "success" or "failure" label to every sampled "first-step action." We chose to analyze the "first-step action" because it is the foundational decision that sets the tone for the entire resolution maneuver and is most representative of the agent's underlying resolution logic.

For visualization, the heatmaps presented in our analysis are generated exclusively from the action samples that were validated as "successful." To clearly represent the three-dimensional hybrid action space, we adopt a "stratify-and-flatten" approach. First, all "successful" action samples are categorized into three distinct groups based on the discrete vertical maneuver command (climb, level flight, or descend). Then, for each group, we generate a 2D histogram heatmap of the planar action distribution (i.e., the change in heading versus the change in speed). By analyzing the number, position, and shape of the peaks in these "successful strategy distribution" heatmaps, we can intuitively and effectively assess the diversity and efficacy of the model's decision-making.

**5.4.2 Experiment Scenarios Design**

To systematically evaluate the model's performance in complex decision-making environments, we designed three representative tactical conflict scenarios. These scenarios were meticulously calibrated with fine-grained constraints to create a decision-making environment that is both challenging and permits the coexistence of multiple resolution pathways, thereby preventing the model from being forced into a single solution by overly restrictive conditions. Asymmetric constraints were intentionally introduced to stimulate the model to generate and exhibit its strategic diversity. In all scenarios, the agent is initialized at flight level FL1 in the center of the airspace, and its action space is constrained by realistic operational rules. The termination conditions for an episode are: 1) reaching a pre-defined waypoint without triggering a Loss of Separation (LoS), or 2) experiencing an LoS before reaching the waypoint. The design of three scenarios is shown in **Fig. 8**.

As shown in **Fig. 8(a)**, the first scenario is a head-on conflict under asymmetric constraints. In this setup, the agent is at FL1 in the center of the airspace with an initial heading of 90°. At FL1, intruder A approaches head-on from right to left with a heading of 270°, while intruder B is positioned to the agent's right with a heading of 80°. At FL2, intruder C is located to the agent's front-left with a heading of 180°. At FL0, intruder D is to the agent's rear-right with a heading of 45°. This configuration introduces asymmetric interference in both the horizontal and vertical dimensions, breaking the optimality of a purely vertical maneuver and forcing a trade-off among several viable strategies, such as "climb and turn left," "descend and accelerate," or "maintain FL1 and turn left."

The second scenario, as shown in **Fig. 8(b)**, features a crossing conflict with dual vertical and lateral constraints. The agent is at FL1 with a heading of 90°. At the same level, intruder A converges from the right on a 90° crossing track (heading 360°), while intruder B is to the agent's front-left with a heading of 130°. At FL2, intruder C is also to the front-left with a heading of 180°. At FL0, intruder D is to the front-right with a heading of 30°, positioned such that an early descent by the agent would induce an LoS. This setup imposes simultaneous constraints on both the lateral and vertical escape routes, encouraging the policy to form a multimodal distribution across maneuvers like "level turn right," "maintain heading and decelerate," and "climb and turn left."

The third scenario, as shown in **Fig. 8(c)**, presents an overtaking/passing conflict under multiple constraints and is the most complex of the three, designed to test the limits of the model's trade-off analysis and fine-grained control capabilities. The agent, at FL1, is preparing to overtake a slower aircraft, A, directly ahead. At the same level, intruder B is on the right with a heading of 60°, and intruder C is on the left with a heading of 120°. At FL2, intruder D is on the right with a heading of 60°, and at FL0, intruder E is on the left with a heading of 120°. This scenario applies pressure from both horizontal sides and the adjacent vertical layers, concentrating the primary viable solutions on "maintain heading and decelerate," "climb and turn left," and "descend and turn right."

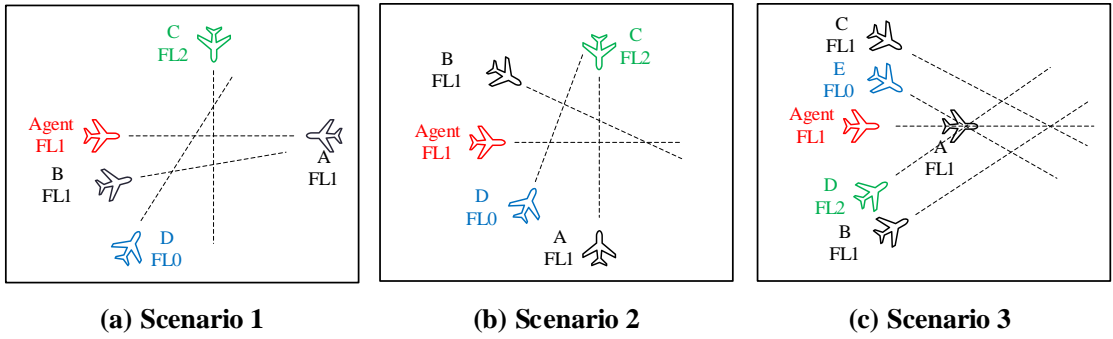

(a) Scenario 1      (b) Scenario 2      (c) Scenario 3

Fig. 8 Multimodal Action Distribution Experiment Scenarios Design

**5.4.3 Experimental Results**

The experimental results for Diffusion-AC and PPO are visualized in the heatmaps shown in **Fig. 9**. Under identical experimental conditions, these visualizations intuitively reveal the fundamental differences in the decision-making patterns of the two algorithms. The results show that, while maintaining a high success rate comparable to that of PPO, Diffusion-AC's initial action distribution consistently forms two or more distinct and stable modal clusters in each scenario. In stark contrast, PPO tends to converge to a single, dominant mode. Specifically, PPO predominantly employs a "climb and turn left" strategy in Scenario 1, relies on a "level flight and turn right" maneuver in Scenario 2, and adopts a "descend and turn right" approach as its absolute core strategy in Scenario 3. Although a few other choices exist outside of these dominant strategies, their probabilities are significantly lower, failing to form stable, alternative decision patterns.

Diffusion-AC, on the other hand, exhibits complementary, high-probability decision regions in all three scenarios. In addition to encompassing a dominant strategy similar to PPO's, it consistently offers effective alternatives such as "level flight, maintain heading, and decelerate." Crucially, these alternative solutions command significant probability mass across different flight levels, demonstrating a structurally consistent, cross-level multimodal characteristic rather than an incidental random distribution.

From the perspective of flight dynamics and operational constraints, the multimodal distributions formed by Diffusion-AC are not merely a "diffused unimodal peak" but correspond to distinct categories of resolution logic. One class of strategies prioritizes composite maneuvers, rapidly establishing separation by simultaneously changing altitude and heading. Another class focuses on speed management, increasing the safety margin by adjusting speed to avoid unnecessary vertical deviations. This strategic distribution aligns perfectly with the design of the three scenarios, as well as with realistic airspace rules and aircraft performance envelopes. Herein lies the core difference from PPO: those suboptimal-but-equally-viable strategic options are fully preserved in Diffusion-AC, where they form clearly discernible modes.

This divergence in decision-making patterns stems from the intrinsic training mechanisms of the two algorithms. Diffusion-AC generates its policy using a value-guided soft teacher distribution and the reverse process of a diffusion model. This dual mechanism ensures the primacy of high-value decision regions while simultaneously preventing premature policy collapse to a "winner-takes-all" solution through temperature and noise scheduling, thereby preserving the probability mass of multiple viable solutions. Conversely, the unimodal nature of the standard PPO algorithm is inherently prone to "mode collapse," causing alternative feasible solutions to appear only sporadically in the long tail of the probability distribution.

In conclusion, Diffusion-AC, without sacrificing safety or success, provides an "interpretable and operational" library of diverse initial decision options. This allows the system, in a real-world deployment, to flexibly switch between multiple equivalent solutions based on real-time constraints, thereby significantly enhancing its robustness to environmental perturbations and execution deviations. This conclusion is jointly substantiated by the consistent emergence of multimodal heatmaps across all three scenarios and the clear, alternative strategic patterns they reveal.

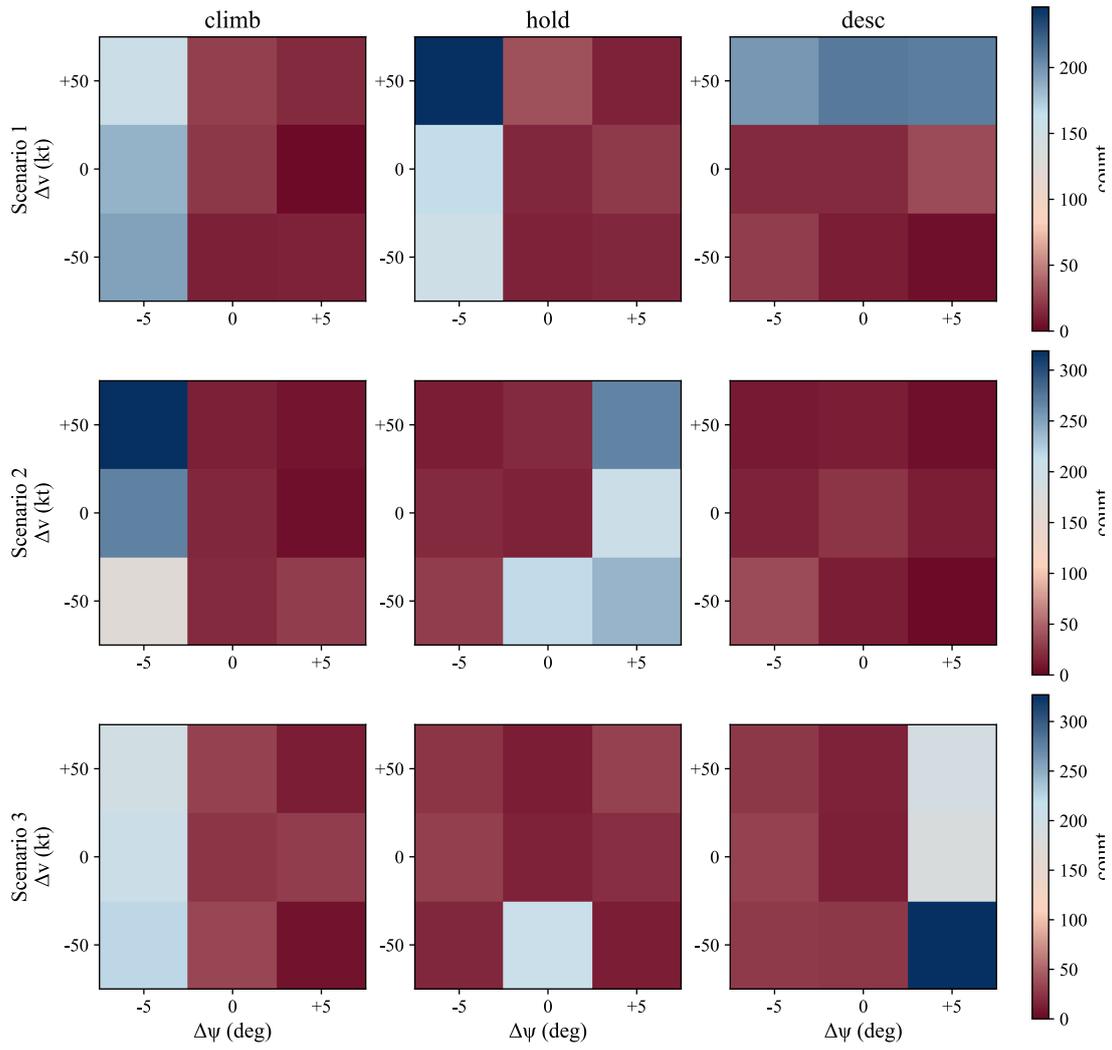

**(a) Diffusion-AC**

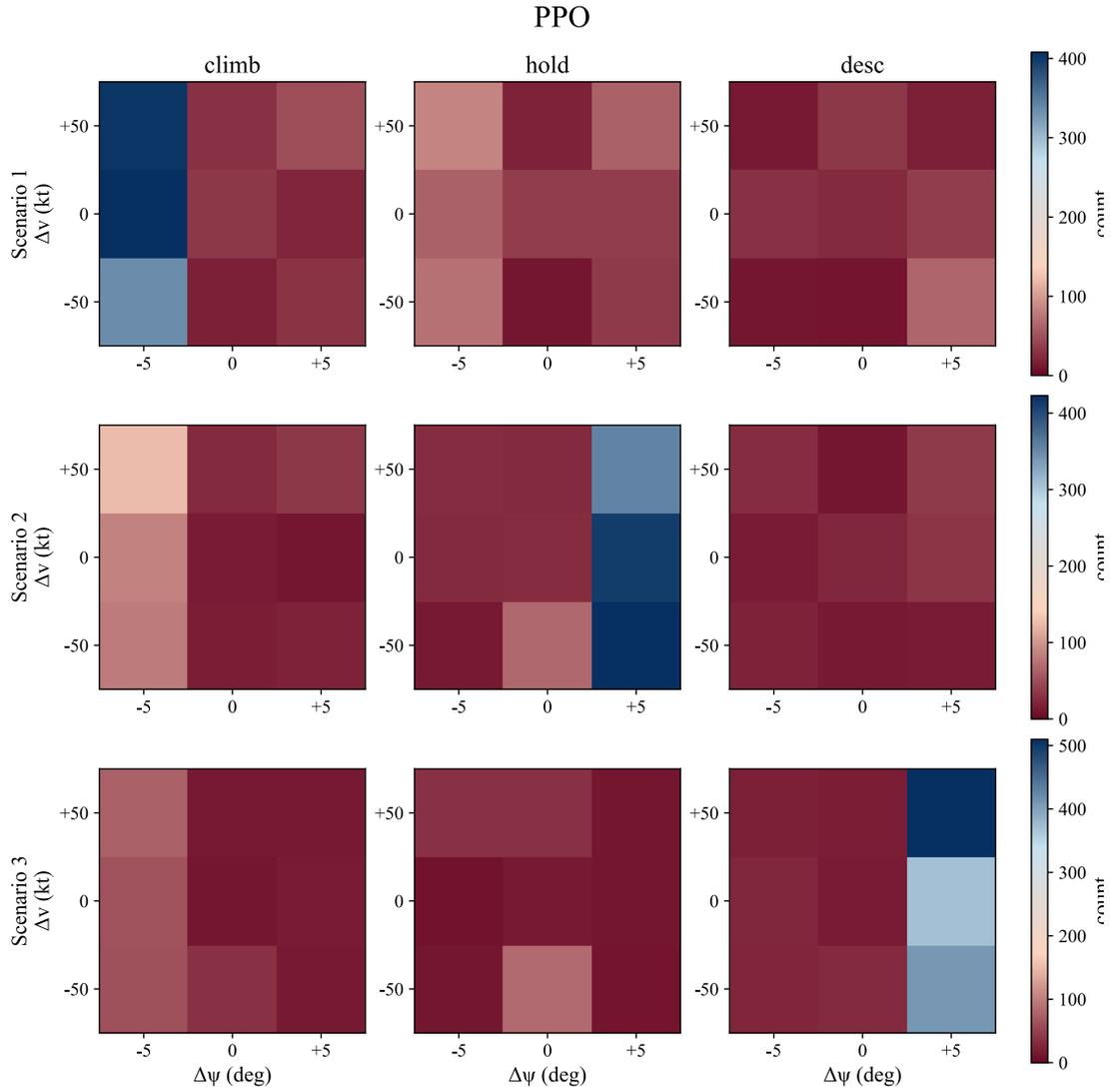

(b) PPO

**Fig. 9 Multimodal Action Distribution Experiment Result Display**

**5.4.4 Experiment Summary**

Through a series of visualized, comparative experiments in three meticulously constrained and complex tactical scenarios, this section has quantitatively and intuitively validated the core advantage of the Diffusion-AC model. The results clearly reveal a fundamental difference in the decision-making patterns of the two algorithms: when faced with the same conflict, the policy distribution of the standard PPO algorithm tends to collapse to a single, dominant solution. In contrast, our proposed Diffusion-AC consistently generates a multimodal distribution comprising several distinct, separated, and empirically validated resolution strategies. These coexisting policy peaks are not simple random perturbations but correspond to solutions with different underlying logic. This finding provides compelling evidence that Diffusion-AC successfully overcomes the mode collapse problem inherent in traditional reinforcement learning algorithms. By preserving "suboptimal-but-equally-viable" options, it endows the autonomous decision-making system with unprecedented flexibility and robustness.

# 6 Conclusion

This paper addressed a fundamental challenge in contemporary deep reinforcement learning approaches for air traffic Conflict Detection and Resolution (CD&R): the inherent unimodal bias of policies, which leads to a critical lack of decision-making flexibility. To overcome this limitation, we introduced Diffusion-AC, a novel autonomous conflict resolution framework that, for the first time, integrates diffusion probabilistic models into a safety-critical decision-making task. Diverging from conventional methods that converge to a single optimal solution, Diffusion-AC models its policy as a reverse denoising process guided by a value function, enabling it to generate a rich, high-quality, and multimodal action distribution. This core architecture is complemented by a Density-Progressive Safety Curriculum (DPSC), a training mechanism that ensures stable and efficient learning as the agent progresses from sparse to high-density traffic environments. The result is an agent capable of retaining and outputting a diverse set of logically sound and equally safe resolution strategies for the same conflict scenario.

Extensive simulation experiments have decisively validated the superior performance of our proposed method. In direct comparisons with a suite of state-of-the-art reinforcement learning benchmarks, Diffusion-AC demonstrated the optimal balance of safety and efficiency across all traffic densities. Most critically, in the most challenging high-density scenarios, our method not only maintained a high success rate of 94.1% but also reduced the incidence of Near Mid-Air Collisions (NMACs) by approximately 59% compared to the next best-performing baseline, significantly enhancing the system's safety margin. This performance leap is a direct consequence of its unique multimodal decision-making capability, which allows the agent to rapidly switch to effective alternative maneuvers when faced with emergent constraints (such as adverse weather) or asymmetric conflicts, thereby avoiding the "decision deadlocks" inherent in unimodal policies. In conclusion, this research not only provides a powerful and substantially more robust solution to the complex CD&R problem but also validates the immense potential of diffusion models as a highly expressive policy representation tool for safety-critical autonomous systems. By doing so, it opens up a new theoretical and practical pathway for the development of the next generation of intelligent air traffic management systems.

# References

[1] Federal Aviation Administration, "2018-2019 NextGen Implementation Plan," Washington DC,USA, 2019.

[2] ICAO, "Working Document for the Aviation System Block Upgrades–The Framework for Global Harmonization," Montreal,Canada, 2013.

[3] SESAR, "European ATM Master Plan – The Roadmap for Delivering High Performing Aviation for Europe, Edition 2015," Brussels, Belgium, 2015.

[4] Brittain, M. W., and Wei, P., "One to Any: Distributed Conflict Resolution with Deep Multi-Agent Reinforcement Learning and Long Short-Term Memory," *AIAA Scitech 2021 Forum*, American Institute of Aeronautics and Astronautics, 2021. https://doi.org/10.2514/6.2021-1952

[5] Brittain, M., and Wei, P., "Autonomous Air Traffic Controller: A Deep Multi-Agent Reinforcement Learning Approach." https://doi.org/10.48550/arXiv.1905.01303

[6] Chen, Y., Hu, M., Yang, L., Xu, Y., and Xie, H., "General Multi-Agent Reinforcement Learning Integrating Adaptive Manoeuvre Strategy for Real-Time Multi-Aircraft Conflict Resolution," *Transportation Research Part C: Emerging Technologies*, Vol. 151, 2023, p. 104125. https://doi.org/10.1016/j.trc.2023.104125

[7] Chen, Y., Xu, Y., Yang, L., and Hu, M., "General Real-Time Three-Dimensional Multi-Aircraft Conflict Resolution Method Using Multi-Agent Reinforcement Learning," *Transportation Research Part C: Emerging Technologies*, Vol. 157, 2023, p. 104367. https://doi.org/10.1016/j.trc.2023.104367

[8] Groot, D. J., Ellerbroek, J., and Hoekstra, J. M., "Analysis of the Impact of Traffic Density on Training of Reinforcement Learning Based Conflict Resolution Methods for Drones," *Engineering Applications of Artificial Intelligence*, Vol. 133, 2024, p. 108066. https://doi.org/10.1016/j.engappai.2024.108066

[9] Wang, L., Yang, H., Lin, Y., Yin, S., and Wu, Y., "Enhancing Air Traffic Control: A Transparent Deep Reinforcement Learning Framework for Autonomous Conflict Resolution," *Expert Systems with Applications*, Vol. 260, 2025, p. 125389. https://doi.org/10.1016/j.eswa.2024.125389

[10] Sui, D., Xu, W., and Zhang, K., "Study on the Resolution of Multi-Aircraft Flight Conflicts Based on an IDQN," *Chinese Journal of Aeronautics*, Vol. 35, No. 2, 2022, pp. 195–213. https://doi.org/10.1016/j.cja.2021.03.015

[11] Li, Y., Zhang, Y., Guo, T., Liu, Y., Lv, Y., and Du, W., "Graph Reinforcement Learning for Multi-Aircraft Conflict Resolution," *IEEE Transactions on Intelligent Vehicles*, Vol. 9, No. 3, 2024, pp. 4529–4540. https://doi.org/10.1109/TIV.2024.3364652

[12] Papadopoulos, G., Bastas, A., Vouros, G. A., Crook, I., Andrienko, N., Andrienko, G., and Cordero, J. M., "Deep Reinforcement Learning in Service of Air Traffic Controllers to Resolve Tactical Conflicts," *Expert Systems with Applications*, Vol. 236, 2024, p. 121234. https://doi.org/10.1016/j.eswa.2023.121234

[13] Kuchar, J. K., and Yang, L. C., "A Review of Conflict Detection and Resolution Modeling Methods," *IEEE Transactions on Intelligent Transportation Systems*, Vol. 1, No. 4, 2000, pp. 179–189. https://doi.org/10.1109/6979.898217

[14] Tomlin, C., Pappas, G. J., and Sastry, S., "Conflict Resolution for Air Traffic Management: A Study in Multiagent Hybrid Systems," *IEEE Transactions on Automatic Control*, Vol. 43, No. 4, 1998, pp. 509–521. https://doi.org/10.1109/9.664154

[15] Mitchell, I. M., Bayen, A. M., and Tomlin, C. J., "A Time-Dependent Hamilton-Jacobi

Formulation of Reachable Sets for Continuous Dynamic Games," *IEEE Transactions on Automatic Control*, Vol. 50, No. 7, 2005, pp. 947–957. https://doi.org/10.1109/TAC.2005.851439

[16] Heinz Erzberger, "Automated Conflict Resolution for Air Traffic Control," NASA Ames Research Center, NASA Ames Research Center Moffett Field, CA, United States, 2005.

[17] Heinz Erzberger, Todd A Lauderdale, and Yung-Cheng Chu, "Automated Conflict Resolution, Arrival Management and Weather Avoidance for ATM," NASA Ames Research Center, NASA Ames Research Center Moffett Field, CA, United States.

[18] Farley, T., Kupfer, M., and Erzberger, H., "Automated Conflict Resolution: A Simulation Evaluation Under High Demand Including Merging Arrivals," *7th AIAA ATIO Conf, 2nd CEIAT Int'l Conf on Innov and Integr in Aero Sciences,17th LTA Systems Tech Conf; followed by 2nd TEOS Forum*, American Institute of Aeronautics and Astronautics, 2007. https://doi.org/10.2514/6.2007-7736

[19] Pallottino, L., Feron, E. M., and Bicchi, A., "Conflict Resolution Problems for Air Traffic Management Systems Solved with Mixed Integer Programming," *Trans. Intell. Transport. Syst.*, Vol. 3, No. 1, 2002, pp. 3–11. https://doi.org/10.1109/6979.994791

[20] van den Berg, J., Guy, S., Lin, M., and Manocha, D., "Reciprocal N-Body Collision Avoidance," *Springer Tracts in Advanced Robotics*, Vol. 70, 2011, pp. 3–19. https://doi.org/10.1007/978-3-642-19457-3_1

[21] Hoekstra, J. M., and Ellerbroek, J., "BlueSky ATC Simulator Project: An Open Data and Open Source Approach."

[22] Groot, D. J., Leto, G., Vlaskin, A., Moec, A., and Ellerbroek, J., "BlueSky-Gym: Reinforcement Learning Environments for Air Traffic Applications."

[23] Brittain, M., Yang, X., and Wei, P., "A Deep Multi-Agent Reinforcement Learning Approach to Autonomous Separation Assurance." https://doi.org/10.48550/arXiv.2003.08353

[24] Brittain, M. W., Alvarez, L. E., and Breeden, K., "Improving Autonomous Separation Assurance through Distributed Reinforcement Learning with Attention Networks," *Proceedings of the AAAI Conference on Artificial Intelligence*, Vol. 38, No. 21, 2024, pp. 22857–22863. https://doi.org/10.1609/aaai.v38i21.30321

[25] Vouros, G., Papadopoulos, G., Bastas, A., Cordero, J. M., and Rodrigez, R. R., "Automating the Resolution of Flight Conflicts: Deep Reinforcement Learning in Service of Air Traffic Controllers." https://doi.org/10.48550/arXiv.2206.07403

[26] Sui, D., Ma, C., and Dong, J., "Conflict Resolution Strategy Based on Deep Reinforcement Learning for Air Traffic Management," *Aviation*, Vol. 27, No. 3, 2023, pp. 177–186. https://doi.org/10.3846/aviation.2023.19720

[27] Wang, Z., Pan, W., Li, H., Wang, X., and Zuo, Q., "Review of Deep Reinforcement Learning Approaches for Conflict Resolution in Air Traffic Control," *Aerospace*, Vol. 9, No. 6, 2022, p. 294. https://doi.org/10.3390/aerospace9060294

[28] Huang, X., Tian, Y., Li, J., Zhang, N., Dong, X., Lv, Y., and Li, Z., "Joint Autonomous Decision-Making of Conflict Resolution and Aircraft Scheduling Based on Triple-Aspect Improved Multi-Agent Reinforcement Learning," *Expert Systems with Applications*, Vol. 275, 2025, p. 127024. https://doi.org/10.1016/j.eswa.2025.127024

[29] Isufaj, R., Aranega Sebastia, D., and Angel Piera, M., "Toward Conflict Resolution with Deep Multi-Agent Reinforcement Learning," *Journal of Air Transportation*, Vol. 30, No. 3, 2022, pp. 71–80. https://doi.org/10.2514/1.D0296

[30]     Zhao, P., and Liu, Y., "Physics Informed Deep Reinforcement Learning for Aircraft Conflict Resolution," *IEEE Transactions on Intelligent Transportation Systems*, Vol. 23, No. 7, 2022, pp. 8288–8301. https://doi.org/10.1109/TITS.2021.3077572

[31]     Sui, D., Ma, C., and Wei, C., "Tactical Conflict Solver Assisting Air Traffic Controllers Using Deep Reinforcement Learning," *Aerospace*, Vol. 10, No. 2, 2023, p. 182. https://doi.org/10.3390/aerospace10020182

[32]     Nilsson, J., Unger, J., and Eilertsen, G., "Self-Prioritizing Multi-Agent Reinforcement Learning for Conflict Resolution in Air Traffic Control with Limited Instructions," *Aerospace*, Vol. 12, No. 2, 2025, p. 88. https://doi.org/10.3390/aerospace12020088

[33]     Han, Y., and Huang, X., "Autonomous Air Traffic Separation Assurance through Machine Learning," *Journal of Industrial and Management Optimization*, Vol. 20, No. 10, 2024, pp. 3195–3204. https://doi.org/10.3934/jimo.2024050

[34]     Deniz, S., and Wang, Z., "Autonomous Conflict Resolution in Urban Air Mobility: A Deep Multi-Agent Reinforcement Learning Approach," *AIAA AVIATION FORUM AND ASCEND 2024*, American Institute of Aeronautics and Astronautics, 2024. https://doi.org/10.2514/6.2024-4005

[35]     Xu, Q., Chen, Z., Li, F., Shen, Z., and Wei, W., "An Efficient Aircraft Conflict Detection and Resolution Method Based on an Improved Reinforcement Learning Framework," *International Journal of Aerospace Engineering*, Vol. 2023, 2023, pp. 1–16. https://doi.org/10.1155/2023/6643903

[36]     Brittain, M., "Autonomous Aircraft Sequencing and Separation with Hierarchical Deep Reinforcement Learning," 2018.

[37]     Yang, L., Huang, Z., Lei, F., Zhong, Y., Yang, Y., Fang, C., Wen, S., Zhou, B., and Lin, Z., "Policy Representation via Diffusion Probability Model for Reinforcement Learning." https://doi.org/10.48550/arXiv.2305.13122

[38]     Wang, Z., Hunt, J. J., and Zhou, M., "Diffusion Policies as an Expressive Policy Class for Offline Reinforcement Learning." https://doi.org/10.48550/arXiv.2208.06193

[39]     Chi, C., Xu, Z., Feng, S., Cousineau, E., Du, Y., Burchfiel, B., Tedrake, R., and Song, S., "Diffusion Policy: Visuomotor Policy Learning via Action Diffusion." https://doi.org/10.48550/arXiv.2303.04137

[40]     Ajay, A., Du, Y., Gupta, A., Tenenbaum, J., Jaakkola, T., and Agrawal, P., "Is Conditional Generative Modeling All You Need for Decision-Making?" https://doi.org/10.48550/arXiv.2211.15657

[41]     Kang, B., Ma, X., Du, C., Pang, T., and Yan, S., "Efficient Diffusion Policies For Offline Reinforcement Learning," Vol. 36, edited by A. Oh, T. Naumann, A. Globerson, K. Saenko, M. Hardt, and S. Levine, 2023, pp. 67195–67212.

[42]     Jackson, M. T., Matthews, M. T., Lu, C., Ellis, B., Whiteson, S., and Foerster, J., "Policy-Guided Diffusion." https://doi.org/10.48550/arXiv.2404.06356

[43]     Ding, S., Hu, K., Zhang, Z., Ren, K., Zhang, W., Yu, J., Wang, J., and Shi, Y., "Diffusion-Based Reinforcement Learning via Q-Weighted Variational Policy Optimization." https://doi.org/10.48550/arXiv.2405.16173

[44]     Luo, K., XIAO, C., Huang, Z., Ling, Z., Fang, Y., and Su, H., "DreamFuser: Value-Guided Diffusion Policy for Offline Reinforcement Learning," 2024.

[45]     He, N., Li, S., Li, Z., Liu, Y., and He, Y., "ReDiffuser: Reliable Decision-Making Using a Diffuserwith Confidence Estimation."

[46]  Zhu, Z., Zhao, H., He, H., Zhong, Y., Zhang, S., Yu, Y., and Zhang, W., "Diffusion Models for Reinforcement Learning: A Survey," *arXiv preprint arXiv:2311.01223*, 2023.

[47]  Balloch, J. C., Bhagat, R., Zollicoffer, G., Jia, R., Kim, J., and Riedl, M. O., "Is Exploration All You Need? Effective Exploration Characteristics for Transfer in Reinforcement Learning." https://doi.org/10.48550/arXiv.2404.02235

[48]  Kushwaha, A., Ravish, K., Lamba, P., and Kumar, P., "A Survey of Safe Reinforcement Learning and Constrained MDPs: A Technical Survey on Single-Agent and Multi-Agent Safety." https://doi.org/10.48550/arXiv.2505.17342

[49]  Schulman, J., Wolski, F., Dhariwal, P., Radford, A., and Klimov, O., "Proximal Policy Optimization Algorithms." https://doi.org/10.48550/arXiv.1707.06347

[50]  Fujimoto, S., Hoof, H. van, and Meger, D., "Addressing Function Approximation Error in Actor-Critic Methods." https://doi.org/10.48550/arXiv.1802.09477